\documentclass[sigconf]{acmart}
\AtBeginDocument{%
  }

\copyrightyear{2026}
\acmYear{2026}
\setcopyright{cc}
\setcctype{by}
\acmConference[WWW '26]{Proceedings of the ACM Web Conference 2026}{April 13--17, 2026}{Dubai, United Arab Emirates}
\acmBooktitle{Proceedings of the ACM Web Conference 2026 (WWW '26), April 13--17, 2026, Dubai, United Arab Emirates}
\acmPrice{}
\acmDOI{10.1145/3774904.3792352}
\acmISBN{979-8-4007-2307-0/2026/04}

\usepackage{xcolor}
\usepackage{hyperref} 
\definecolor{blueblack}{rgb}{0.1,0.2,0.5} 
\usepackage{multirow}       
\usepackage[normalem]{ulem} 
\useunder{\uline}{\ul}{}   
\usepackage{array}          
\usepackage{makecell}       
\usepackage{subcaption}     
\usepackage{caption}        

\usepackage{pifont}         
\usepackage{colortbl}       

\usepackage{enumitem}

\newcommand{\better}[1]{\textsubscript{\scriptsize\textcolor{red!70!black}{$\blacktriangledown$\,#1}}}
\newcolumntype{L}[1]{>{\raggedright\arraybackslash}m{#1}}

\newcommand{\best}[1]{\textcolor{red}{\textbf{#1}}}
\newcommand{\second}[1]{\textcolor{blue}{#1}}

\begin{document}

\title{HALO: Hierarchical Reinforcement Learning for Large-Scale Adaptive Traffic Signal Control}


\author{Yaqiao Zhu}
\email{yz1034@exeter.ac.uk}
\orcid{0009-0004-1482-3240}
\affiliation{
  \institution{University of Exeter}
  \department{Department of Computer Science}
  \city{Exeter}
  \country{UK}
}

\author{Hongkai Wen}
\email{hongkai.wen@warwick.ac.uk}
\orcid{0000-0003-1159-090X}
\affiliation{
  \institution{University of Warwick}
  \department{Department of Computer Science}
  \city{Coventry}
  \country{UK}
}

\author{Geyong Min}
\email{g.min@exeter.ac.uk}
\orcid{0000-0003-1395-7314}
\affiliation{
  \institution{University of Exeter}
  \department{Department of Computer Science}
  \city{Exeter}
  \country{UK}
}

\author{Man Luo}
\authornote{Corresponding author.}
\email{m.luo@exeter.ac.uk}
\orcid{0000-0002-7346-9024}
\affiliation{
  \institution{University of Exeter}
  \department{Department of Computer Science}
  \city{Exeter}
  \country{UK}
}

\begin{abstract}
  Adaptive traffic signal control (ATSC) is essential for mitigating urban congestion in modern smart cities, 
  where traffic infrastructure is evolving into interconnected Web-of-Things (WoT) environments with thousands of sensing-and-control nodes. 
  However, existing methods face a critical scalability-coordination tradeoff: centralized approaches optimize global objectives but become computationally intractable at city scale, while decentralized multi-agent methods scale efficiently yet lack network-level coherence, resulting in suboptimal performance.
  In this paper, we present HALO, a hierarchical reinforcement learning framework that addresses this tradeoff for large-scale ATSC. 
  HALO decouples decision-making into two levels: a high-level global guidance policy employs Transformer-LSTM encoders to model spatio-temporal dependencies across the entire network and broadcast compact guidance signals, while low-level local intersection policies execute decentralized control conditioned on both local observations and global context.  
  To ensure better alignment of global-local objectives, we introduce an adversarial goal-setting mechanism where the global policy proposes challenging-yet-feasible network-level targets that local policies are trained to surpass, fostering robust coordination. 
  We evaluate HALO extensively on multiple standard benchmarks, and a newly constructed large-scale Manhattan-like network with 2,668 intersections under real-world traffic patterns, including peak transitions, adverse weather and holiday surges. Results demonstrate HALO shows competitive performance and becomes increasingly dominant as network complexity grows across small-scale benchmarks, while delivering the strongest performance in all large-scale regimes, offering up to 6.8\% lower average travel time and 5.0\% lower average delay than the best state-of-the-art.

\end{abstract}



\maketitle
\newcommand\webconfavailabilityurl{https://doi.org/10.1145/3774904.3792352}
\ifdefempty{\webconfavailabilityurl}{}{
\begingroup\small\noindent\raggedright\textbf{Resource Availability:}\\
The source code of this paper has been made publicly available at \url{https://github.com/Julietjobs/HALO}.
\endgroup
}

\section{Introduction}

In modern smart cities, traffic infrastructure is rapidly evolving into a dense Web-of-Things (WoT) environment. Intersections are no longer isolated points with traffic lights; they have become interconnected sensing-and-control nodes in a citywide web, transforming urban road networks into an integrated, city-scale system. This paradigm shift presents a tangible opportunity for globally coordinated traffic management, where adaptive traffic signal control (ATSC) - systems that continuously adjust signal behaviour (e.g., cycle length, splits and offsets) in response to real-time data (e.g., from inductive loop detectors, roadside cameras, and connected vehicles) to reduce delay, queues, and stops - is increasingly deployed in practice. However, in real-world operation ATSC unfolds within a tightly coupled web of cause and effect: a small timing change at one intersection can reshape platoons, trigger queue spillback, or fracture green waves several corridors away. These interactions are non-linear, potentially delayed, and fundamentally non-local, so consequences often emerge far from where they originate. Optimizing such a dynamic, interactive system clearly exceeds the scope of the traditional static, rule-based approaches~\cite{roess2004traffic,kouvelas2014maximum}, which typically fail to account for emergent, system-wide dynamics.

\begin{figure}[tbp]
    \begin{subcaptionbox}{Centralized vs. Multi-agent RL\label{RLmethods}}[0.55\linewidth]
        {\includegraphics[width=\linewidth]{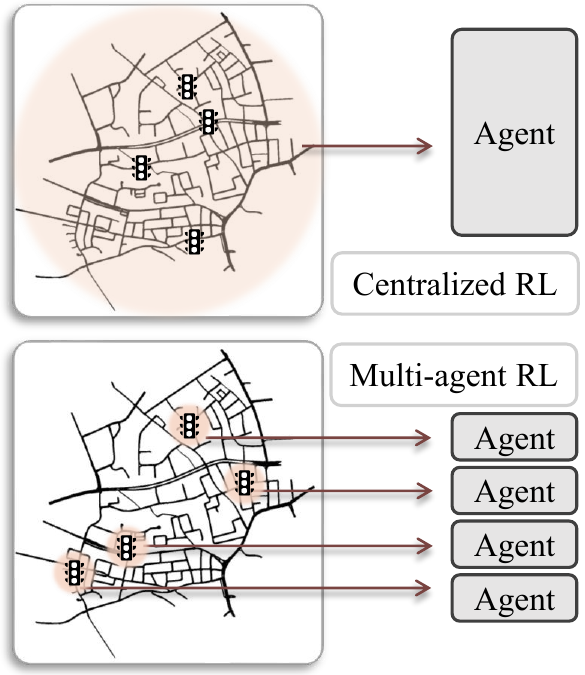}}
    \end{subcaptionbox}
    \hfill
    \begin{subcaptionbox}{HALO framework\label{HALO_framework}}[0.37\linewidth]
        {\includegraphics[width=\linewidth]{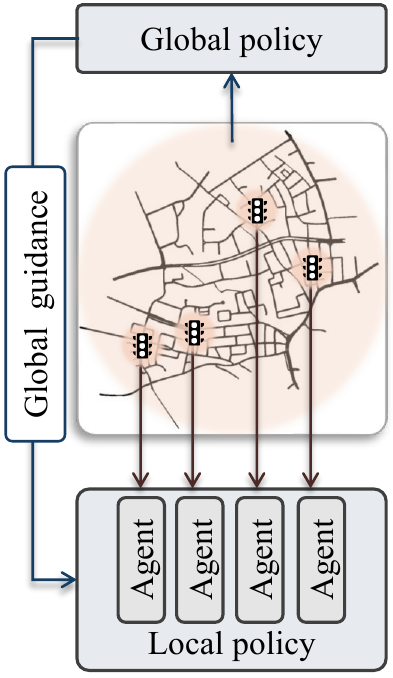}}
    \end{subcaptionbox}
    \caption{(a) Traditional centralized vs. decentralized (multi-agnet) reinforcement learning (RL) paradigms for ATSC. (b) The proposed HALO framework, which uses a global policy to provide guidance to decentralized local policies.}
    \vspace{-1\baselineskip}
    \Description{This figure compares different reinforcement learning paradigms for adaptive traffic signal control.
    The left part illustrates two existing approaches: centralized and multi-agent reinforcement learning.
    In the centralized scheme, a single agent controls all intersections in the network, enabling global optimization but facing scalability issues.
    In the multi-agent scheme, each intersection has its own agent, which improves scalability but may lead to uncoordinated local behaviors.
    The right part shows the proposed HALO hierarchical framework.
    It introduces a global policy that periodically broadcasts compact global guidance to multiple local agents, each controlling one intersection through its local policy.
    The bidirectional arrows indicate information flow between the global and local levels, combining top-down coordination with bottom-up feedback.
    Together, the figure highlights how HALO balances scalability and global coherence in large-scale adaptive traffic signal control.
    }
\end{figure}

Recently, Reinforcement Learning (RL) has become a promising paradigm for ATSC, which aims to learn a policy that maps the observed traffic states (e.g., queues/waiting times, current signal phase) to signal actions (e.g., phase changes, green splits, cycle length/offset), so that a certain reward function (e.g., negative delay/queues/stops, or throughput) is maximized~\cite{DBLP:conf/ijcai/MeiLSW23,du2024felight,gu2024pi}. Broadly speaking, RL-based ATSC approaches can be categorized into two classes, depending on how control decisions are coordinated, as shown in Figure~\ref{RLmethods}. \textit{Centralized RL} based ATSC approaches typically employ a single agent controlling all intersections over the entire road network, and therefore optimize a global objective - often can discover globally optimal policy but become computationally intractable as the joint state-action space explodes. On the other hand, \textit{decentralized RL} (multi-agent RL) based ATSC methods assign an agent to each intersection, improving scalability but risking local myopia - agents may stuck in policies that lack coordination among the others, leading to suboptimal decisions that hinder overall performance across the entire road network. To mitigate this, recent multi-agent RL (MARL) based ATSC approaches typically adopt different forms of communication and/or coordination techniques, e.g., graph-attention messaging ~\cite{wei2019colight}, 
pressure-based parameter sharing~\cite{chen2020toward}, non-local attention~\cite{lin2023denselight}, and dynamic grouping~\cite{liu2023gplight} to balance coordination and local decisions between agents. Despite successes in certain cases, those approaches still face key limitations when considered in the real world: (i) scalability is bought at the cost of global coherence, where communication and coordination between agents (e.g., neighbor-only messaging or clustering of groups) are rather flat and only happen regionally, leaving global dynamics under-modeled; (ii) agents often struggle to account for effects that emerge far from, and/or long after certain actions being made - they do not have explict mechnisms to capture the non-local, long-range dependencies which are common in real-world traffic conditions; (iii) most of them are trained with limited synthetic benchmarks or rather simplified configurations, and thus learned policies are often brittle under real-world dynamics, e.g., peak hours, incidents, holiday rush or advsese weather. 

To address these challenges, in this paper we propose HALO, a novel hierarchical reinforcement learning framework with adversarial goal-setting for large-scale adaptive traffic signal control. At its core, HALO separates \textit{what the entire road network should achieve} from \textit{how each road intersection should act}: it is designed to directly tackle the scalability-coordination trade-off through a two-level architecture, as shown in Figure~\ref{HALO_framework}. It learns a \textit{global guidance policy} and \textit{local intersection policy} respectively, where the former acts as a high-level controller that processes aggregated features of traffic over the whole road network and periodically broadcasts guidance vectors indicating global priorities (e.g., congestion due to incidents), while the latter is per-junction policy that selects signal actions from their local observations, conditioned on the received global guidance. Training is end-to-end with a mixed global–local objective (using a centralized critic during learning), while execution is fully decentralized apart from the lightweight broadcast of guidance vectors. More importantly, HALO considers an adversarial goal setting mechanism, where the global guidance policy is rewarded for proposing challenging-but-feasible goals, while local policies are trained to beat those goals - turning the global–local coupling into a minimax game that avoids trivial solutions and fosters more robust alignment between local executions and global coordination at scale. More concretely, our technical contributions are as follows:
\begin{itemize}[leftmargin=6mm]
    \item We propose HALO, a novel hierachical RL framework to solve the critical trade-off between scalability and coordination in large-scale ATSC, which decouples the city-scale objectives (via a global guidance policy) and local decisions (via local intersection policies) in an organic way, so that high-level traffic dynamics and context can be injected into the decentralized control of traffic signals, enabling agents to make locally adaptive decisions with citywide awareness.
    \item We introduce a novel joint optimization mechanism for both global and local policies via adversarial goal setting. We let the global guidance policy learn to propose challenging but feasible goals, while local intersection policies are trained to exceed them - enhancing coordination between agents while improving sample efficiency under sparse rewards. 
    \item We conduct an extensive evaluation of HALO on both standard benchmarks and a large-scale Manhattan-like setting (containing 2668 intersections), with three real-world traffic patterns (peak transition, adverse weather, holiday rush). Results show that HALO outperforms state-of-the-art MARL-based ATSC baselines, offering up to 6.8\% and 5.0\% improvements in average trip time and delays respectively, with significant gains under realistic flows in large-scale road networks.
\end{itemize}

\section{Related Work}

\noindent\textbf{Adaptive Traffic Signal Control (ATSC).} 
ATSC dynamically adjusts traffic light behavior to improve traffic flow across a road network. Traditional methods~\cite{roess2004traffic,kouvelas2014maximum,sims1980sydney}rely on heuristics or local feedback and struggle under non-stationary, uncertain demand. Earlier learning-based work - kernel methods and single-agent reinforcement learning~\cite{aslani2017adaptive,chu2014kernel,richter2006natural,genders2018evaluating,genders2016using} adopt centralized control paradigm: therefore effective on small networks but failing to scale because the joint state–action space grows exponentially with the number of intersections.

\noindent \textbf {Multi-agent RL (MARL) for ATSC.} 
MARL-based ATSC approaches assign one agent per intersection. Many approaches adopt centralized training with decentralized execution - causing agents to behave uncoordinatedly at inference, so local gains can degrade global performance~\cite{da2024prompt,da2023uncertainty,ault2019learning,yu2022surprising}. To encourage coordination, recent work adds explicit collaboration, e.g. neighbor messaging with graph neural networks~\cite{wang2021adaptive,zhang2020generalight}. However, these mechanisms operate only within limited regions and do not provide citywide context, leaving a mismatch between local optimization and global objectives.

\noindent \textbf{ATSC for Large-Scale Road Networks.} Recently there has been a growing interest in pushing ATSC for city-scale road networks - where scalability and global coordination must be achieved simultaneously. For instance, MPLight~\cite{chen2020toward} uses pressure-based rewards with parameter sharing, demonstrating feasibility at high intersection counts. To capture longer-range dependencies, DenseLight~\cite{lin2023denselight} applies non-local attention with low-rank decomposition, while GPLight~\cite{liu2023gplight} forms dynamically clustered groups for agents. Beyond a single city, X-Light, FedLight, and MetaLight/MetaVIM employ meta-/federated learning to improve cross-scenario transfer~\cite{DBLP:conf/ijcai/Jiang00XRLM024,ye2021fedlight,zang2020metalight,zhu2023metavim}. However these approaches either trade citywide coordination for efficiency (limited by neighborhood or group-bounded coordination), or incur prohibitive computation/communication costs to model interactions across the whole road networks.

\noindent\textbf{Hierarchical Reinforcement Learning (HRL).} 
HRL approaches typically structure policies across multiple levels, so that high-level policies set goals while low-level ones execute them~\cite{sutton1999between, bacon2017option,andrychowicz2017hindsight,nachum2018data,vezhnevets2017feudal, DBLP:conf/iclr/FransH0AS18}. This separation improves exploration and captures long-range dependencies, which are essential in the ATSC context. However, most of the existing HRL-based ATSC approaches either: (i) employ a centralized manager that issues sub-goals to all interactions - capturing some global structure but losing scalability~\cite{zeng2022halight}; or (ii) learn per-intersection actions without a shared citywide signal - scalable but still locally myopic and sensitive to action design~\cite{xu2021hierarchically, yang2023hierarchical}. Moreover, hand-crafted sub-goal spaces often cannot cope well with the diverse, time-varying traffic patterns in the real world. In contrast, our approach adopts a globally guided, decentralized HRL formulation: a lightweight global guidance signal conditions all local controllers, and adversarial goal setting encourages informative, non-trivial solutions robust to real-world traffic dynamics.

\begin{figure}[t]
  \centering
  \includegraphics[width=0.78\linewidth]{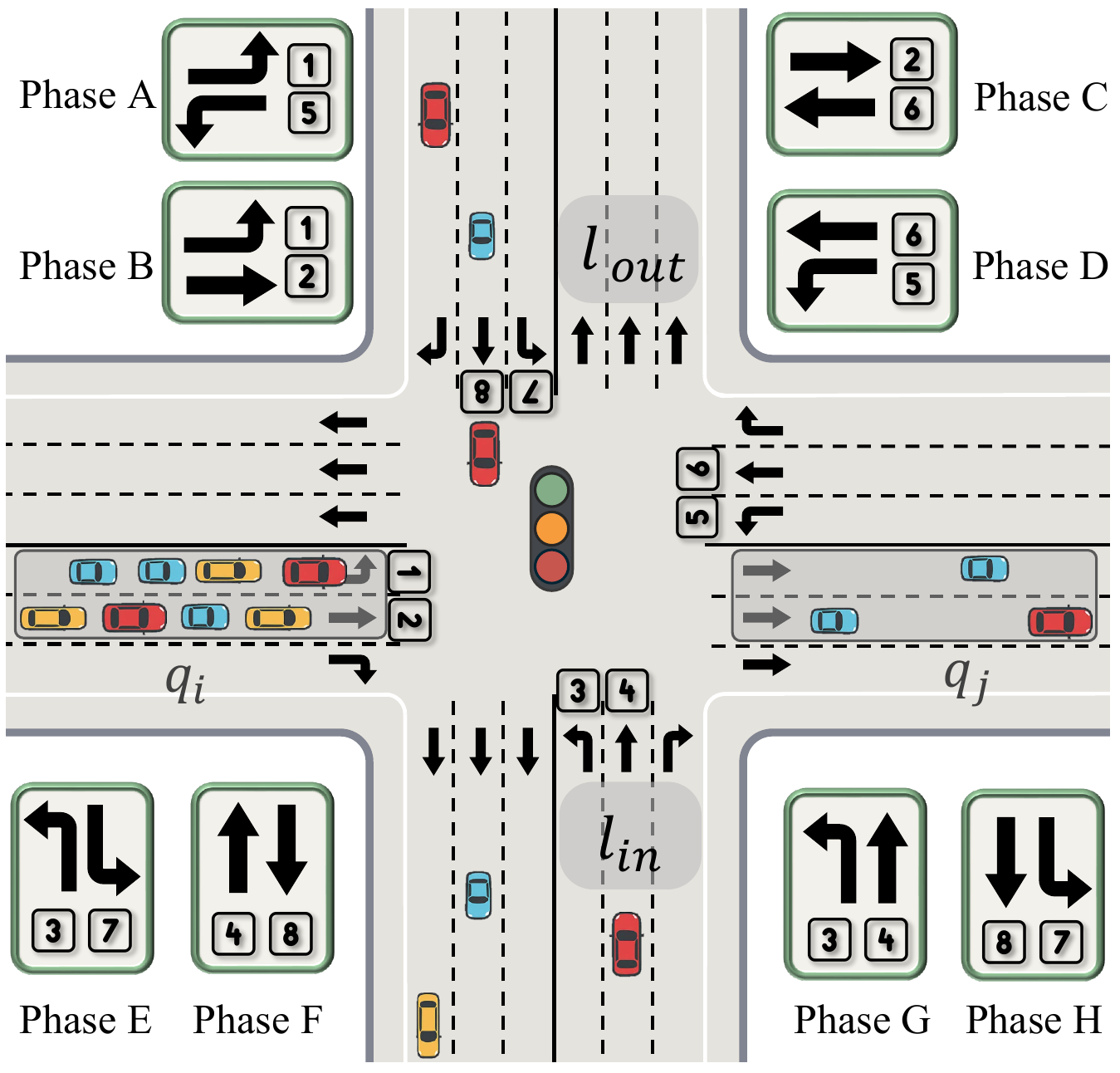}
  \vspace{-1\baselineskip}
  \caption{Four-leg signalized intersection with eight non-conflicting phases (A–H).}
  \vspace{-1\baselineskip}
  \label{figure2}
  \Description{
This figure illustrates a standard four-leg signalized road intersection with lane configurations and eight signal phases labelled A to H.
Each approach has three incoming lanes and three outgoing lanes representing left-turn, through, and right-turn movements.
For each movement, traffic arriving from an incoming lane lin with queue length qi proceeds to its corresponding outgoing lane lout with queue length qj.
The difference qi minus qj represents the pressure of that movement.
Each signal phase, such as Phase A or Phase E, combines several non-conflicting movements that can run simultaneously.
The total intersection pressure is obtained by summing all individual movement pressures across active lanes.
Colored cars and arrows illustrate vehicle flows for different phases, while the central traffic light marks the intersection control point.
This diagram clarifies how pressure and phases are defined for subsequent reinforcement-learning-based control models.
}

\end{figure}

\section{Problem Definition}
\label{definition}

\subsection{Preliminaries}

\noindent \textbf{Traffic Intersections.} 
An intersection refers to a location in the road network where two or more roads cross or merge, typically equipped with traffic signals to regulate vehicle movement. Figure~\ref{figure2} shows a typical example: a standard four-leg intersection (also called a four-arm intersection), where four roads meet, forming a cross shape. Each approach (leg) contains three incoming lanes $l_{in}$ dedicated to left-turn, through, and right-turn movements, and typically three corresponding outgoing lanes $l_{out}$.

\noindent \textbf{Signal Phase.}
A signal phase is the set of non‑conflicting movements that receive right‑of‑way simultaneously at an intersection. As illustrated in Figure~\ref{figure2}, a standard intersection typically adopts eight phases, each governing vehicle movements such as left turns, through movements, and right turns. In some scenarios, a single incoming lane may accommodate multiple movement directions. For example, in the Arterial 4×4 scenario, certain lanes serve both through and right-turning vehicles. Additionally, in accordance with real-world practices, right-turning vehicles are not regulated by traffic signals.

\noindent \textbf{Pressure.}
Pressure is a metric used to quantify the imbalance of traffic flow at an intersection, specifically between incoming and outgoing lanes. It is typically defined as the difference between the queue lengths (or vehicle densities) of incoming and outgoing lanes for a specific movement. The total pressure $P$ for a phase is usually calculated as the sum of pressures over all movements controlled by that phase: $P = \sum_{(i,j)} (q_i - q_j)$, where $q_i$ is the number of vehicles on the incoming lane $l_{in}^{i}$, and $q_j$ is that on the corresponding outgoing lane $l_{out}^{j}$, as shown in Figure~\ref{figure2}.

\subsection{Problem Formulation}

We formulate the large-scale ATSC problem as a Hierarchical Multi-agent Markov Decision Process (H-MMDP), defined by the tuple $\langle \mathcal{I}, \mathcal{S}, \mathcal{O}, \mathcal{A}, \mathcal{P}, \mathcal{R}, \gamma \rangle$, where $\mathcal{I} = \{1, 2, \dots, N\}$ is the set of $N$ intersections, each controlled by a local intersection policy agent with shared parameters. $\mathcal{S}$ denotes the global state space, capturing the complete traffic configuration across the network. $\mathcal{O} = \{o_i\}_{i \in \mathcal{I}}$ represents the set of local observations, where each $o_i$ includes intersection $i$'s local traffic state, neighboring information, and global guidance from the global policy. $\mathcal{A} = \mathcal{A}^H \times \mathcal{A}^L$ is the hierarchical action space, where $\mathcal{A}^H$ is the global policy's action space (generating global sub-goals), and $\mathcal{A}^L = \prod_{i \in \mathcal{I}} \mathcal{A}_i^L$ is the joint local action space with each $\mathcal{A}_i^L$ representing signal phase selections at intersection $i$. $\mathcal{P}: \mathcal{S} \times \mathcal{A} \rightarrow \Delta(\mathcal{S})$ is the state transition function. $\mathcal{R}: \mathcal{S} \times \mathcal{A} \rightarrow \mathbb{R}$ is the reward function combining global and local objectives. $\gamma \in [0,1)$ is the discount factor.

The joint policy $\pi$ is factorized into two levels that operate at different temporal and spatial scales. At the global level, the global policy observes aggregated network state $s \in \mathcal{S}$ and periodically generates a global guidance signal $g^t = (F_g, G^t)$, where $F_g$ is a compact embedding providing contextual awareness and $G^t = (G_w^t, G_q^t)$ specifies measurable network-level targets for waiting time and queue length. At the intersection level, each agent $i$ observes its local state $o_i$ (including local traffic conditions, neighbor information, and the global guidance $g^t$) and selects a signal action $a_i^L \in \mathcal{A}_i^L$. All local intersection policy agents share the same network parameters but act based on their respective local observations. The factorized joint policy can be expressed as:
\begin{equation}
\pi(a^H, \{a_i^L\}_{i \in \mathcal{I}} \mid s) = \pi^H(g^t \mid s) \cdot \prod_{i=1}^N \pi^L(a_i^L \mid o_i, g^t)
\end{equation}

The reward function $\mathcal{R}$ integrates both global coordination and local efficiency:

\begin{equation}
\mathcal{R}(s, g^t, \{a_i^L\}_{i \in \mathcal{I}}, s') = \mathcal{R}^H(s, g^t, s') + \sum_{i \in \mathcal{I}} \mathcal{R}_i^L(o_i, a_i^L, o_i')
\end{equation}
where $\mathcal{R}^H$ evaluates the alignment between achieved network-level performance and the global policy's targets $G^t$, and $\mathcal{R}_i^L$ measures local traffic efficiency (e.g., queue length, waiting time, pressure) at intersection $i$. The goal is to learn a joint policy that maximizes the expected cumulative reward:
\begin{equation}
\max_{\pi} \; \mathbb{E}_{\pi} \left[ \sum_{t=0}^{\infty} \gamma^t \mathcal{R}(s_t, g^t, \{a_{i,t}^L\}_{i \in \mathcal{I}}, s_{t+1}) \right]
\end{equation}

This hierarchical formulation enables HALO to maintain global coordination through the global policy's guidance signals while preserving scalability through decentralized local intersection policies execution.

\section{Methodology}
\label{methods}

Large-scale ATSC requires both (i) global coordination across distant intersections and (ii) local responsiveness to rapidly changing queues. HALO resolves this by decoupling “global planning” and “local execution”: a \emph{global guidance policy} summarizes network-wide dynamics and sets a compact global goal, while decentralized \emph{local intersection policies} make per-intersection decisions conditioned on both local observations and the global guidance. 

\begin{figure*}
    \centering
    \includegraphics[width=\linewidth]{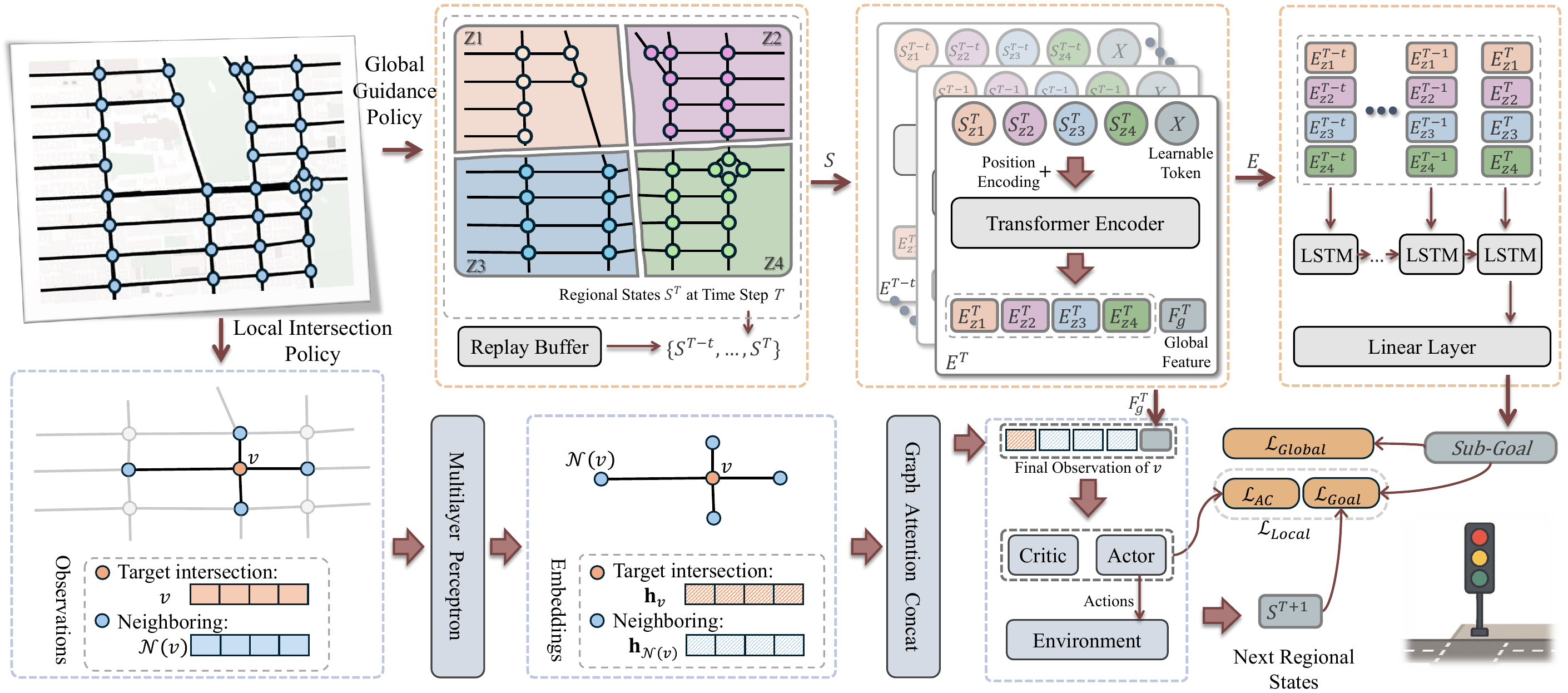}
    \vspace{-2\baselineskip}
    \caption{An overview of the two-level HALO pipeline. The global policy (yellow) compresses regional states with a Transformer and LSTMs to produce a broadcast guidance $F_g$ and a measurable sub-goal $G^t$; the local intersection policy (blue) forms per-junction observations from the target node and its nearest neighbors, applies MLP + direction-aware graph attention, appends $F_g$ for Actor–Critic control; adversarial losses $\mathcal L_{\text{Global}}, \mathcal L_{\text{AC}}, \mathcal L_{\text{Goal}}$ couple the two levels and drive the next-state feedback.}
    \vspace{-1\baselineskip}
    \label{figure1}
    \Description{
    This figure illustrates the complete workflow of the HALO hierarchical reinforcement learning framework for large-scale adaptive traffic signal control. 
    The diagram is divided into two color-coded parts: the yellow dashed box represents the global policy and the blue dashed box represents the local intersection policies. 
    On the left, the global policy receives regional traffic states from the entire city network, aggregates them into subregions, and stores historical states in a replay buffer. 
    A Transformer encoder with a learnable global token captures spatial correlations among regions, while LSTM layers model temporal evolution to produce a global feature and a measurable sub-goal that specifies network-level targets for waiting time and queue length. 
    On the right, the local intersection policies focuses on individual intersections. Each intersection (target node) observes its local and neighboring traffic conditions, encodes them with a multilayer perceptron, and applies direction-aware graph attention to weigh the influence of nearby intersections. 
    The resulting embeddings are concatenated with the global guidance vector from the global policy to form the final observation. 
    This observation is processed by an Actor–Critic network to generate signal control actions. 
    The actions update the traffic environment, producing new regional states for the next step. 
    Adversarial objectives couple the global policy and local intersection policies so that global goals and local behaviors are aligned, enabling coordinated yet scalable control across the network.
    }
\end{figure*}

As shown in Figure~\ref{figure1}, the high-level global policy employs Transformer and LSTM modules to capture spatiotemporal patterns of the entire network and generates compact global goals, while a shared-parameter local intersection policy makes real-time intersection-level decisions by combining local states with this global guidance. To enhance efficiency under sparse rewards and multi-agent objectives, an adversarial training mechanism is introduced in which the global policy sets progressively more challenging goals and the local intersection policy adapts to outperform them, thereby achieving globally consistent yet locally adaptive control.

\subsection{Global Guidance Policy}

Purely local collaboration often misses long-range dependencies, while fully centralized attention across thousands of intersections is computationally prohibitive. We therefore (i) aggregate intersections into $M$ subregions to keep global modeling compact, (ii) capture spatial coupling \emph{across subregions} per time step via a Transformer with a learnable global token, and (iii) model temporal evolution with an LSTM. The global guidance policy outputs two signals: a global guidance embedding $F_g$ fed to all intersections to enhance context awareness, and a measurable global target $G^t=(G_w^t,G_q^t)$ that specifies desired network-level congestion and waiting levels (i.e. Sub-Goal in Figure~\ref{figure1}).

\subsubsection{Input Representation}
We partition the network into $M$ subregions $\mathcal{Z}=\{z_1,\dots,z_M\}$. At time $t$, each subregion $z$ produces a regional feature vector $\mathbf{s}_z^t\in\mathbb{R}^{d_{\mathrm{reg}}}$ that summarizes its local traffic status and spatial context. Concretely, the regional state aggregates: (i) intensity signals reflecting demand and congestion, such as the total number of stopped vehicles and cumulative waiting time within the subregion; (ii) throughput- and density-related indicators, including the total vehicle count and average lane occupancy aggregated over the subregion’s intersections and incoming lanes; and (iii) spatial coordinates that encode the subregion’s location in the city grid, enabling the model to correlate geography with recurrent congestion patterns. Over a temporal window of length $T$, we form $\mathbf{S}_z=[\mathbf{s}_z^1,\ldots,\mathbf{s}_z^T]\in\mathbb{R}^{T\times d_{\mathrm{reg}}}$ for each $z$, preserving short-term dynamics needed by the global guidance policy for forecasting and target setting.

\subsubsection{Transformer Encoding Across Subregions}
We apply spatial self-attention per time step to exchange information across subregions while keeping the temporal axis outside the Transformer. For each $t$, we prepend a learnable global token $\mathbf{x}_{\mathrm{cls}}\in\mathbb{R}^{d_{\mathrm{reg}}}$ and form
\begin{equation}
\mathbf{X}^{t}=[\mathbf{x}_{\mathrm{cls}},\mathbf{s}_1^{t},\dots,\mathbf{s}_M^{t}]\in\mathbb{R}^{(M+1)\times d_{\mathrm{reg}}}.
\end{equation}

We add sinusoidal positional encodings over subregion indices to preserve spatial order, obtaining $\mathbf{X}^{t}_{\mathrm{pos}}$. The Transformer encoder yields
\begin{equation}
[\mathbf{E}_{G}^{t}, \mathbf{E}_{1}^{t}, \ldots, \mathbf{E}_{M}^{t}] = \mathrm{TransformerEncoder}(\mathbf{X}^{t}_{\mathrm{pos}}).
\end{equation}

Here $\mathbf{E}_{G}^{t}$ summarizes the global spatial context at time $t$ (global token), while $\mathbf{E}_{z}^{t}$ encodes subregion $z$ with cross-region attention. We define the global guidance embedding as $F_g \triangleq \mathbf{E}_{G}^{T}$ at the current step and broadcast it to all intersections as part of their observations.

\subsubsection{LSTM-based Sub-goal Generation}
To provide \emph{measurable} global targets that align local decisions, we model temporal evolution for each subregion and then aggregate to produce a network-level target. We first arrange subregion embeddings across the window:
\begin{equation}
\hat{\mathbf{E}}_z = [\mathbf{E}_z^1, \ldots, \mathbf{E}_z^T] \in \mathbb{R}^{T \times d_{\mathrm{reg}}},\quad z=1,\dots,M.
\end{equation}

Each $\hat{\mathbf{E}}_z$ is passed to an LSTM, yielding a final hidden state $\mathbf{h}_z^T\in\mathbb{R}^{d_{\mathrm{hidden}}}$. We aggregate subregions and project to a 2-D global target:
\begin{equation}
\mathbf{u}=\Big[\mathbf{h}_1^T \,\|\, \cdots \,\|\, \mathbf{h}_M^T\Big],\quad
G^t=\big(G_w^t,G_q^t\big)=\psi(\mathbf{u})\in\mathbb{R}^2,
\end{equation}
where $\psi(\cdot)$ is a lightweight MLP, $\, \| \,$ represents concatenation. $G^t$ serves as the network-level target for total waiting time and total queue length at step $t$, complementing the representation $F_g$. We decouple their roles: $F_g$ enhances decision context at intersections; $G^t$ shapes the global objective for training and evaluation.

\subsection{Local Intersection Policy}
Global guidance policy provides both global representations and high-level optimization objectives, while the core responsibility of the local intersection policy is to perform real-time adaptive traffic signal control decisions at each intersection based on comprehensive observation information. Intersections must react to local conditions but stay aligned with global congestion relief. We therefore fuse (i) locally encoded states with (ii) dynamically weighted nearest-neighbor features and (iii) the global guidance $F_g$ from the global guidance policy. This preserves directionality and responsiveness while injecting compact global context.

Each intersection is modeled as an independent agent, and all agents share the same network parameters. The local intersection policy integrates local traffic observations from the surrounding neighbors with the global guidance representation provided by the global guidance policy, thereby constructing a final observation with multi-scale information awareness. This final observation serves as the input to both the policy and value networks within an Actor-Critic framework.

\subsubsection{Local Observation Encoding}
At each time step, the local observation for a given intersection is composed of its own state and the states of its four nearest neighboring intersections, resulting in a total of five observation vectors. The information from neighboring intersections is dynamically weighted according to its relative importance at the current time step. This design captures both the temporal variability and directional sensitivity of traffic flow. For instance, during morning or evening rush hours, different neighbors may serve as upstream intersections that carry major incoming traffic.

Let $\mathbf{o}_v^t \in \mathbb{R}^k$ denote the raw $k$-dimensional observation vector of intersection $v$ at time $t$. Let $\mathcal{N}(v)$ denote the set of its four nearest neighboring intersections. At time $t$, each observation 
$\mathbf{o}_v^t$ is first encoded into an intermediate representation $\mathbf{h}_v^t$ via a shared Multilayer Perceptron (MLP). The specific observation features are detailed in the Appendix~\ref{app:additional-design-details}

\subsubsection{Dynamic Neighbor Aggregation}
Intersections must respond to rapidly changing local conditions while avoiding myopic behaviors that propagate congestion. Simple averaging of neighbor messages over-smooths directional signals (e.g., dominant upstream inflow), while dense long-range message passing amplifies noise and incurs high cost at city scale. To balance efficiency and sensitivity, we adopt a lightweight, direction-aware neighbor aggregation that learns attention weights over a fixed set of nearest neighbors and concatenates the weighted features, preserving directional information and mitigating over-smoothing.

Let $\mathbf{H}^t=[\mathbf{h}_1^t,\ldots,\mathbf{h}_N^t]\in\mathbb{R}^{N\times F}$ be the encoded features at time $t$ and let $\mathbf{A}\in\mathbb{R}^{N\times N}$ denote the adjacency where $A_{ij}=1$ if node $j\in\mathcal{N}(i)$, with $\mathcal{N}(i)$ the four nearest neighbors measured along the road network. For each valid edge $(i,j)$ with $A_{ij}=1$, we compute an attention score:
\begin{equation}
e_{ij}=\mathrm{LeakyReLU}\!\left(\mathbf{a}^{\top}\,[\mathbf{h}_i^t \,\|\, \mathbf{h}_j^t]\right),
\quad
\alpha_{ij}=\mathrm{softmax}_{j\in\mathcal{N}(i)}(e_{ij})
\end{equation}
where $\mathbf{a}\in\mathbb{R}^{2F}$ is a learnable vector and $\,\|\, $ denotes concatenation. The final neighbor-augmented representation concatenates the self feature with directionally weighted neighbor features:
\begin{equation}
\mathbf{z}_i^t \;=\; \mathbf{h}_i^t \,\|\, \alpha_{ij_1}\mathbf{h}_{j_1}^t \,\|\, \alpha_{ij_2}\mathbf{h}_{j_2}^t \,\|\, \alpha_{ij_3}\mathbf{h}_{j_3}^t \,\|\, \alpha_{ij_4}\mathbf{h}_{j_4}^t \;\in\; \mathbb{R}^{5F}.
\end{equation}

If fewer than four neighbors exist, we zero‑pad to keep a fixed width. Finally, we attach the global policy’s guidance embedding by forming $\mathbf{o}_i^{\mathrm{final},t}=\mathbf{z}_i^t\,\|\,F_g$, which is fed to the actor and critic to enable globally consistent yet locally responsive control.


\begin{table*}[tbp]
\begin{center}
\centering
\renewcommand{\arraystretch}{1.2}
\vspace{-0.5\baselineskip}
\caption{Characteristics of evaluation scenarios. Left: 5 small-scale benchmarks; Right: a large-scale Manhattan-like scenario.}
\vspace{-1\baselineskip}
\label{tbl:settings}

\small
\begin{tabular}{>{\centering\arraybackslash}m{1.8cm}|>{\centering\arraybackslash}m{1.5cm}>{\centering\arraybackslash}m{1.4cm}>{\centering\arraybackslash}m{1.5cm}>{\centering\arraybackslash}m{1.6cm}>{\centering\arraybackslash}m{1.6cm}|>{\centering\arraybackslash}m{1.7cm}>{\centering\arraybackslash}m{1.8cm}>{\centering\arraybackslash}m{1.6cm}}

\noalign{\hrule height 1.5pt}

\cellcolor{gray!20} & 
\cellcolor{gray!20} & 
\cellcolor{gray!20} & 
\cellcolor{gray!20} & 
\cellcolor{gray!20} & 
\cellcolor{gray!20} & 
\multicolumn{3}{c}{\cellcolor{gray!20}\textbf{Manhattan2668}} \\
\cline{7-9}
\cellcolor{gray!20}\multirow{-2}{1.6cm}{\centering\textbf{Scenarios}} & 
\cellcolor{gray!20}\multirow{-2}{1.4cm}{\centering Cologne8} & 
\cellcolor{gray!20}\multirow{-2}{1.4cm}{\centering Grid4×4} & 
\cellcolor{gray!20}\multirow{-2}{1.5cm}{\centering Arterial4×4} & 
\cellcolor{gray!20}\multirow{-2}{1.6cm}{\centering Ingolstadt21} & 
\cellcolor{gray!20}\multirow{-2}{1.4cm}{\centering Grid5×5} & 
\cellcolor{gray!20}Config 1 & 
\cellcolor{gray!20}Config 2 & 
\cellcolor{gray!20}Config 3 \\

\hline

Country & Germany & Synthetic & Synthetic & Germany & Synthetic & United States & United States & United States \\
\cellcolor{black!5}\# Intersections & \cellcolor{black!5}8 & \cellcolor{black!5}16 & \cellcolor{black!5}16 & \cellcolor{black!5}21 & \cellcolor{black!5}25 & \cellcolor{black!5}2668 & \cellcolor{black!5}2668 & \cellcolor{black!5}2668 \\
Min flow (/s) & 0.037 & 0.018 & 0.026 & 0.039 & 0.033 & 1.389 & 1.944 & 4.167 \\
\cellcolor{black!5}Max flow (/s) & \cellcolor{black!5}0.059 & \cellcolor{black!5}0.038 & \cellcolor{black!5}0.185 & \cellcolor{black!5}0.209 & \cellcolor{black!5}0.379 & \cellcolor{black!5}6.944 & \cellcolor{black!5}5.139 & \cellcolor{black!5}7.639 \\
\hline
Flow type & \begin{tabular}[c]{@{}c@{}}Morning peak\\ Real flow\end{tabular} & \begin{tabular}[c]{@{}c@{}}Multi-modal\\ Gaussian\end{tabular} & \begin{tabular}[c]{@{}c@{}}Multi-modal\\ Gaussian\end{tabular} & \begin{tabular}[c]{@{}c@{}}Morning peak\\ Real flow\end{tabular} & \begin{tabular}[c]{@{}c@{}}Multi-modal\\ Gaussian\end{tabular} & \begin{tabular}[c]{@{}c@{}}Peak Transition\\ Real flow\end{tabular} & \begin{tabular}[c]{@{}c@{}}Adverse Weather\\ Real flow\end{tabular} & \begin{tabular}[c]{@{}c@{}}Holiday Rush\\ Real flow\end{tabular} \\

\noalign{\hrule height 1.5pt}

\end{tabular}
\end{center}
\vspace{-1\baselineskip}
\end{table*}

\subsection{Joint Optimization}

In this section, we describe the joint optimization mechanism integrating global guidance policy and local intersection policy under an adversarial hierarchical reinforcement learning framework. We elaborate on reward definitions, trajectory formulation, and adversarial loss functions.

\subsubsection{Reward and Trajectory}
Each intersection $i$ receives a local reward at time $t$:
\begin{equation}
r_i^t \;=\; - \big( ql_i^t + wt_i^t + dt_i^t + ps_i^t \big) \;+\; ss_i^t,
\end{equation}
where $ql, wt, dt, ps, ss$ denote queue length, waiting time, delay time, pressure, and speed score, respectively. The global guidance policy provides a measurable global target $G^t=(G_w^t, G_q^t)$ for the total waiting time $W_{\mathrm{global}}^t$ and total queue length $Q_{\mathrm{global}}^t$ at step $t$. To encourage surpassing the target while keeping the signal well-behaved, we adopt a margin-based hinge formulation:
\begin{equation}
r_g^t \;=\; - \Big[ \beta_w \, (W_{\mathrm{global}}^t - G_w^t)_+ \;+\; \beta_q \, (Q_{\mathrm{global}}^t - G_q^t)_+ \Big],
\end{equation}
where $(x)_+=\max(0,x)$ and $\beta_w,\beta_q>0$ weight the two global metrics. The per-step total reward is $r^t=\sum_i r_i^t + r_g^t$. For intersection $i$, a trajectory element is
\begin{equation}
\tau_i^t \;=\; \big(\mathbf{o}_i^{\mathrm{final},t},\, a_i^t,\, r^t,\, \mathbf{o}_i^{\mathrm{final},t+1}\big),
\end{equation}
and trajectories are formed over the same rollout window used by the local intersection policy’s Actor–Critic updates, ensuring consistent evaluation of both global and local objectives.

\subsubsection{Adversarial Training Mechanism}
The global guidance policy predicts the global evolution and sets slightly ambitious targets to steer the network beyond its current capability. Let $Y^t=(W_{\mathrm{global}}^t, Q_{\mathrm{global}}^t)$ denote the realized global status and $\hat{Y}^t=(\hat{W}^t,\hat{Q}^t)$ the global guidance policy’s prediction over the window. With a non-negative margin $\delta\ge 0$, we form an \emph{ambitious} target by
\begin{equation}
G^t \;=\; \max\!\big(\hat{Y}^t - \delta,\, \mathbf{0}\big)\quad\text{(element-wise)}.
\end{equation}

The global guidance policy minimizes
\begin{equation}
\mathcal{L}_{\mathrm{Global}} \;=\; \mathbb{E}\!\left[ \,\big\| \hat{Y}^t - Y^t \big\|_2^2 \;+\; \eta_1 \, \big\| G^t - (\hat{Y}^t - \delta) \big\|_2^2 \,\right],
\end{equation}
where the first term fits global dynamics and the second maintains an explicit margin-based ambition; $\eta_1>0$ balances them.

The local intersection policy maximizes returns via PPO-style Actor–Critic while aligning its achieved performance with $G^t$ using a convex penalty:
\begin{equation}
\mathcal{L}_{\mathrm{Local}} \;=\; \mathbb{E}\!\left[ \mathcal{L}_{\mathrm{AC}} \;+\; \eta_2 \big( \beta_w \, (W_{\mathrm{global}}^t - G_w^t)_+ + \beta_q \, (Q_{\mathrm{global}}^t - G_q^t)_+ \big) \right],
\end{equation}
with $\eta_2>0$. We adopt two-time-scale updates (global guidance policy updated less frequently than local intersection policy), apply stop-gradient to $G^t$ when updating the local intersection policy, and use gradient clipping and reward normalization. These choices reduce adverse feedback loops and stabilize training under the adversarial coupling while preserving the intended global–local alignment. We include the details of methods used to improve training stability in Appendix~\ref{app:impl-training-details}.

\section{Evaluation}
\label{experiments}

\subsection{Experimental Setup}
\label{setup}
We use the Simulation of Urban MObility (SUMO) traffic simulator~\cite{Krajzewicz2012SUMO} for all experiments. Each episode is set to 3600 seconds of simulated time, with a rollout length of 240 steps. The yellow light duration is 5 seconds, and each signal phase iteration lasts 10 seconds. As shown in Table~\ref{tbl:settings}, we consider five small-scale benchmarks (\textit{Cologne8}, \textit{Grid4$\times$4}, \textit{Arterial4$\times$4}, \textit{Ingolstadt21} and \textit{Grid5$\times$5}) widely used by existing work. In addition, we construct a new large-scale scenario based on road network of Manhattan, containing 2668 traffic lights, and refer to this as \textit{Manhattan2668} hereafter. For this scenario, we define the network using OSMWebWizard\footnote{https://sumo.dlr.de/docs/Tutorials/OSMWebWizard.html} and manually refine its intersection details. Traffic flow is generated based on open-source taxi trip data\footnote{https://opendata.cityofnewyork.us/data}, and we select dates associated with adverse weather conditions using Local Climatological Data\footnote{https://www.ncei.noaa.gov/products/land-based-station/local-climatological-data} to simulate the \textit{Adverse Weather Flow}. In addition, we simulate two other realistic traffic patterns based on actual urban traffic conditions: the \textit{Peak Transition Flow}, which represents a sudden influx of vehicles during the transition from off-peak to peak hours, and the \textit{Holiday Rush Flow}, which reflects surges in traffic during holiday periods with increased travel demand. Visualizations of the experimental scenarios (e.g. intersection topologies) are included in Appendix~\ref{app:exp-vis}. The detailed hyper-parameter settings can be found in Appendix~\ref{app:impl-hyperparams}.


\begin{table}[tbp]
\centering
\renewcommand{\arraystretch}{1.2}

\caption{Performance on standard scenarios. Best in bold red, second best in blue, and improvements shown by \better{\%}.}
\vspace{-1\baselineskip}
\label{tbl:standard}

\small
\begin{tabular}{>{\centering\arraybackslash}m{1.6cm}>{\centering\arraybackslash}m{1.6cm}>{\centering\arraybackslash}m{1.9cm}>{\centering\arraybackslash}m{1.8cm}}

\noalign{\hrule height 1.5pt}

\cellcolor{gray!20}\textbf{Methods} & 
\cellcolor{gray!20}\textbf{Arterial4×4} & 
\cellcolor{gray!20}\textbf{Ingolstadt21} & 
\cellcolor{gray!20}\textbf{Grid5×5} \\

\hline
\multicolumn{4}{l}{\cellcolor{gray!10}\textbf{Avg. Trip Time (seconds) $\downarrow$}} \\
\hline

FTC & 828.38\tiny{±8.17} & 319.41\tiny{±24.48} & 550.38\tiny{±8.31} \\
\cellcolor{black!5} MaxPressure & \cellcolor{black!5} 686.12\tiny{±9.57} & \cellcolor{black!5} 375.25\tiny{±2.40} & \cellcolor{black!5} 274.15\tiny{±15.23} \\
CoLight & 409.93\tiny{±0.00} & 337.46\tiny{±0.00} & 242.37\tiny{±0.00} \\
\cellcolor{black!5} MPLight & \cellcolor{black!5} 541.29\tiny{±45.24} & \cellcolor{black!5} 319.28\tiny{±10.48} & \cellcolor{black!5} 261.76\tiny{±6.60} \\
MetaLight & 381.77\tiny{±12.85} & 292.26\tiny{±4.40} & 247.83\tiny{±5.99} \\
\cellcolor{black!5} IPPO & \cellcolor{black!5} 431.31\tiny{±28.55} & \cellcolor{black!5} 379.22\tiny{±34.03} & \cellcolor{black!5} 259.28\tiny{±9.55} \\
rMAPPO & 565.67\tiny{±44.8} & 453.61\tiny{±29.66} & 300.9\tiny{±8.31} \\
\cellcolor{black!5} MetaGAT & \cellcolor{black!5} 374.8\tiny{±0.87} & \cellcolor{black!5} 290.73\tiny{±0.45} & \cellcolor{black!5} 266.6\tiny{±0.00} \\
GESA & 393.57\tiny{±13.72} & 320.02\tiny{±5.57} & 252.11\tiny{±9.94} \\
\cellcolor{black!5} CoSLight & \cellcolor{black!5} 364.21\tiny{±4.78} & \cellcolor{black!5} 284.62\tiny{±2.57} & \cellcolor{black!5} \second{ 220.32\tiny{±1.71}} \\
X-Light & \second{ 349.6\tiny{±0.00}} & \second{ 278.05\tiny{±0.00}} & 220.63\tiny{±0.00} \\
\rowcolor{green!10} \textbf{HALO} & \best{341.42\tiny{±4.86}}\better{2.3\%} & \best{272.50\tiny{±6.03}}\better{2.0\%} & \best{204.26\tiny{±8.14}}\better{7.3\%} \\

\hline
\multicolumn{4}{l}{\cellcolor{gray!10}\textbf{Avg. Delay Time (seconds) $\downarrow$}} \\
\hline

FTC & 1234.30\tiny{±6.50} & 183.70\tiny{±26.21} & 790.18\tiny{±7.96} \\
\cellcolor{black!5} MaxPressure & \cellcolor{black!5} 952.53\tiny{±12.48} & \cellcolor{black!5} 275.36\tiny{±14.38} & \cellcolor{black!5} 240.00\tiny{±18.43} \\
CoLight & 776.61\tiny{±0.00} & 226.06\tiny{±0.00} & 248.32\tiny{±0.00} \\
\cellcolor{black!5} MPLight & \cellcolor{black!5} 1083.18\tiny{±63.38} & \cellcolor{black!5} 185.04\tiny{±10.70} & \cellcolor{black!5} 213.78\tiny{±14.44} \\
MetaLight & 862.32\tiny{±39.01} & 164.80\tiny{±3.75} & 209.13\tiny{±19.40} \\
\cellcolor{black!5} IPPO & \cellcolor{black!5} 914.58\tiny{±36.90} & \cellcolor{black!5} 247.68\tiny{±35.33} & \cellcolor{black!5} 243.58\tiny{±9.29} \\
rMAPPO & 1185.20\tiny{±167.48} & 372.20\tiny{±39.85} & 346.78\tiny{±28.25} \\
\cellcolor{black!5} MetaGAT & \cellcolor{black!5} 772.36\tiny{±0.00} & \cellcolor{black!5} 176.86\tiny{±2.37} & \cellcolor{black!5} 234.8\tiny{±0.00} \\
GESA & 775.22\tiny{±8.63} & 209.57\tiny{±3.32} & 210.74\tiny{±13.56} \\
\cellcolor{black!5} CoSLight & \cellcolor{black!5} 744.98\tiny{±16.49} & \cellcolor{black!5} 179.73\tiny{±14.26} & \cellcolor{black!5} \second{ 178.54\tiny{±4.34}} \\
X-Light & \second{697.79\tiny{±0.00}} & \second{ 160.39\tiny{±0.00}} & 187.74\tiny{±0.00} \\
\rowcolor{green!10} \textbf{HALO} & \best{682.86\tiny{±8.95}}\better{2.1\%} & \best{155.70\tiny{±1.56}}\better{2.9\%} & \best{164.92\tiny{±3.70}}\better{7.6\%} \\

\noalign{\hrule height 1.5pt}

\end{tabular}
\vspace{-1\baselineskip}
\end{table}



\begin{table*}[tbp]
\begin{center}
\centering
\vspace{0pt}
\renewcommand{\arraystretch}{1.2}
\vspace{-0.5\baselineskip}
\caption{Performance of competing approaches on Manhattan2668 under three traffic flow settings. Best in bold red, second best in blue, and improvements shown by \better{\%}.}
\vspace{-1\baselineskip}
\label{tbl:manhattan}

\footnotesize
\resizebox{\textwidth}{!}{
\begin{tabular}{>{\centering\arraybackslash}m{1.6cm}|>{\centering\arraybackslash}m{1.6cm}>{\centering\arraybackslash}m{1.8cm}>{\centering\arraybackslash}m{1.6cm}|>{\centering\arraybackslash}m{1.6cm}>{\centering\arraybackslash}m{1.8cm}>{\centering\arraybackslash}m{1.6cm}}

\noalign{\hrule height 1.5pt}

\cellcolor{gray!20} & 
\multicolumn{3}{c|}{\cellcolor{gray!20}\textbf{Avg. Trip Time (seconds) $\downarrow$}} & 
\multicolumn{3}{c}{\cellcolor{gray!20}\textbf{Avg. Delay Time (seconds) $\downarrow$}} \\
\cline{2-4}\cline{5-7}
\cellcolor{gray!20}\multirow{-2}{1.6cm}{\centering\textbf{Methods}} & 
\cellcolor{gray!20}Peak Transition & 
\cellcolor{gray!20}Adverse Weather & 
\cellcolor{gray!20}Holiday Rush & 
\cellcolor{gray!20}Peak Transition & 
\cellcolor{gray!20}Adverse Weather & 
\cellcolor{gray!20}Holiday Rush \\

\hline

FTC & 1474.83\tiny{±11.45} & 1942.79\tiny{±5.14} & 2116.07\tiny{±10.81} & 1385.62\tiny{±24.99} & 1530.38\tiny{±29.6} & 1517.91\tiny{±19.78} \\
\cellcolor{black!5} MaxPressure & \cellcolor{black!5} 1175.25\tiny{±8.35} & \cellcolor{black!5} 1481.05\tiny{±9.34} & \cellcolor{black!5} 1667.4\tiny{±7.09} & \cellcolor{black!5} 975.95\tiny{±14.13} & \cellcolor{black!5} 1092.04\tiny{±11.00} & \cellcolor{black!5} 1133.57\tiny{±18.29} \\
CoLight & 907.43\tiny{±0.00} & 1194.92\tiny{±0.00} & 1313.62\tiny{±0.00} & 713.15\tiny{±0.00} & 793.59\tiny{±0.00} & 783.52\tiny{±0.00} \\
\cellcolor{black!5} MPLight & \cellcolor{black!5} 1111.42\tiny{±23.52} & \cellcolor{black!5} 1463.75\tiny{±10.37} & \cellcolor{black!5} 1549.02\tiny{±21.5} & \cellcolor{black!5} 800.14\tiny{±1.59} & \cellcolor{black!5} 887.25\tiny{±15.4} & \cellcolor{black!5} 907.16\tiny{±7.78} \\
MetaLight & 823.6\tiny{±24.37} & 1085.5\tiny{±89.04} & 1178.03\tiny{±86.22} & \second{ 567.55\tiny{±33.01}} & 752.64\tiny{±34.57} & 746.17\tiny{±40.72} \\
\cellcolor{black!5} IPPO & \cellcolor{black!5} 981.12\tiny{±58.13} & \cellcolor{black!5} 1486.93\tiny{±75.09} & \cellcolor{black!5} 1411.91\tiny{±82.16} & \cellcolor{black!5} 794.88\tiny{±16.84} & \cellcolor{black!5} 926.85\tiny{±14.58} & \cellcolor{black!5} 901.33\tiny{±10.01} \\
rMAPPO & 1140.17\tiny{±35.26} & 1338.17\tiny{±12.07} & 1568.61\tiny{±26.50} & 863.48\tiny{±33.89} & 904.72\tiny{±25.77} & 901.33\tiny{±11.40} \\
\cellcolor{black!5} MetaGAT & \cellcolor{black!5} 772.92\tiny{±0.00} & \cellcolor{black!5} 1024.78\tiny{±0.00} & \cellcolor{black!5} 1127.09\tiny{±0.00} & \cellcolor{black!5} 681.61\tiny{±0.00} & \cellcolor{black!5} 761.8\tiny{±0.00} & \cellcolor{black!5} 756.24\tiny{±0.00} \\
GESA & \second{ 732.28\tiny{±7.52}} & \second{ 930.47\tiny{±18.72}} & 1181.53\tiny{±17.65} & 673.82\tiny{±5.39} & 723.29\tiny{±8.39} & \second{ 629.82\tiny{±7.74}} \\
\cellcolor{black!5} CoSLight & \cellcolor{black!5} 783.15\tiny{±14.96} & \cellcolor{black!5} 1094.27\tiny{±34.30} & \cellcolor{black!5} 1284.27\tiny{±28.87} & \cellcolor{black!5} 586.93\tiny{±12.04} & \cellcolor{black!5} 728.94\tiny{±9.37} & \cellcolor{black!5} 662.32\tiny{±10.94} \\
X-Light & 843.36\tiny{±0.00} & 1103.85\tiny{±0.00} & \second{ 1051.61\tiny{±0.00}} & 634.07\tiny{±0.00} & \second{ 667.22\tiny{±0.00}} & 641.89\tiny{±0.00} \\
\rowcolor{green!10} \textbf{HALO (ours)} & \best{690.88\tiny{±9.01}}\better{5.7\%} & \best{913.84\tiny{±14.9}}\better{1.8\%} & \best{980.24\tiny{±20.44}}\better{6.8\%} & \best{549.02\tiny{±6.24}}\better{3.3\%} & \best{649.71\tiny{±1.40}}\better{2.6\%} & \best{598.65\tiny{±9.16}}\better{5.0\%} \\

\noalign{\hrule height 1.5pt}

\end{tabular}}
\end{center}
\vspace{-1\baselineskip}
\end{table*}

\subsection{Metrics and Baselines}
We employ two widely used evaluation metrics in ATSC: \textit{Average Travel Time} (ATT) and \textit{Average Delay Time} (ADT). ATT measures the average duration that all vehicles spend in the scenario, from entering to exiting. ADT captures the average delay caused by congestion or traffic signals. More concretely, ATT and ADT are defined as follows:
\begin{equation}
{ATT} = \frac{1}{V} \sum_{i=1}^{V} \left(t^l_i - t^e_i\right), {ADT} = \frac{1}{V} \sum_{i=1}^{V} \left(T^r_i - T^t_i\right)
\end{equation}
where $t^e_i$ and $t^l_i $represent the entry and exit times of the $i$-th vehicle. $T^r_i$ denotes the actual travel time of the $i$-th vehicle (i.e., $T^r_i = t^l_i - t^e_i$), and $T^t_i$ represents the theoretical travel time of the $i$-th vehicle assuming it travels at the allowed speed without any delay.

We evaluate HALO against both classical and state-of-the-art MARL-based ATSC approaches: 

\begin{itemize}[leftmargin=6mm]
\item \textbf{FTC}~\cite{roess2004traffic}, a fixed-time, rule-based controller that cycles through a predetermined plan without adapting to real-time conditions;
\item \textbf{MaxPressure}~\cite{kouvelas2014maximum}, which selects the phase with the highest queue length (i.e. pressure) to maximize throughput;
\item \textbf{IPPO}~\cite{ault2021reinforcement}, which runs an independent PPO agent at each intersection with no explicit coordination at inference;
\item \textbf{rMAPPO}~\cite{yu2022surprising}, which also assigns PPO agents per intersection but trains with shared data to encourage coordinated learning;
\item \textbf{CoLight}~\cite{wei2019colight}, which uses graph-attention messaging to incorporate spatial–temporal influence from neighbors;
\item \textbf{MPLight}~\cite{chen2020toward}, which leverages pressure-based RL with parameter sharing to improve performance across large networks;
\item \textbf{MetaLight}~\cite{zang2020metalight}, which applies gradient-based meta-RL to transfer knowledge and adapt quickly to new traffic scenarios;
\item \textbf{MetaGAT}~\cite{lou2022meta}, which combines context-based meta-learning with a graph-attention network to improve generalization;
\item \textbf{GESA}~\cite{jiang2024general}, which considers a scenario-agnostic, unified intersection representations for zero-shot transfer; 
\item \textbf{CoSLight}~\cite{ruan2024coslight}, which learns a collaborator selection policy to pick effective partners beyond immediate neighbors;
\item \textbf{X-Light}~\cite{DBLP:conf/ijcai/Jiang00XRLM024} which uses a dual-level Transformer to aggregate local and cross-city knowledge for better transferability.
\end{itemize}

\subsection{Results}
\label{exp:results}

\begin{figure}[t]
  \centering
  \includegraphics[width=\linewidth]{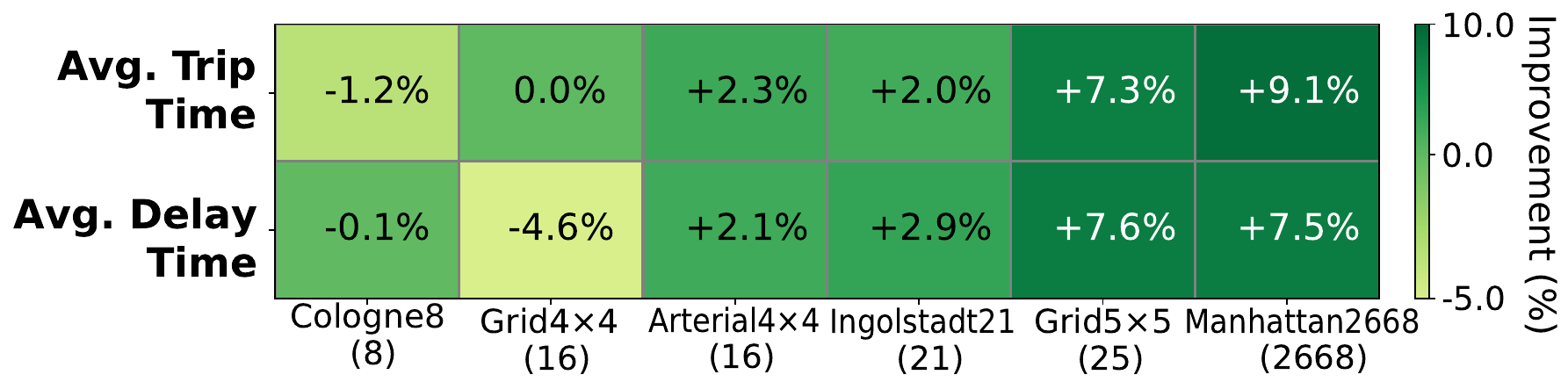}
  \vspace{-1.5\baselineskip}
  \caption{Relative gains (\%) of HALO vs. the strongest baseline ordered by road network sizes (left to right, small to large). Color encodes improvement (darker means better).}
  \vspace{-1\baselineskip}
  \label{fig:gain-vs-size}
  \Description{This heatmap shows HALO's relative performance gains versus the strongest baseline across six scenarios ordered by network size: Cologne8 (8 intersections), Grid4×4 (16), Arterial4×4 (16), Ingolstadt21 (21), Grid5×5 (25), and Manhattan2668 (2,668).
  Two metrics are displayed—Average Trip Time (ATT) and Average Delay Time (ADT)—with darker colors indicating larger improvements.
  A clear scaling trend emerges: minimal gains on small networks (near-zero on Cologne8), modest gains on medium networks (approximately 2\% on Arterial4×4 and Ingolstadt21), and substantial advantages on large networks (7–8\% on Grid5×5, 7–9\% on Manhattan2668).
  This pattern validates that HALO's hierarchical global coordination becomes increasingly valuable as network scale grows, effectively resolving the scalability-coordination trade-off in large-scale traffic control.}
\end{figure}

\noindent\textbf{Performance on Standard Benchmarks.}
As shown in Table~\ref{tbl:standard}, HALO offers the best performance on all three standard benchmarks: Arterial4$\times$4, Ingolstadt21 and Grid5$\times$5, where results on Cologne8 and Grid4$\times$4 are included in Table~\ref{tbl:additional-results} in Appendix~\ref{app:exp-additional-results} due to space limits. We see that HALO is increasingly dominant as the size of road networks grows. On Arterial4$\times$4 (16 intersections), HALO achieves the best ATT (341.42s), a 2.3\% gain over the strongest baseline (X-Light), and exhibits a similar margin on ADT (2.1\% better), and on Ingolstadt21 (21 intersections) it achieves similar performance gains (ATT 2.0\%, ADT 2.9\% better). On Grid5$\times$5 (25 intersections), the advantage widens markedly (ATT 7.3\%, ADT 7.6\% better), indicating that broadcast guidance and adversarial goal setting help coordinate decisions across multiple intersections and mitigate spillback that neighbor-only methods miss. Additional cases (Cologne8 and Grid4$\times$4) are in Table~\ref{tbl:additional-results} (Appendix~\ref{app:exp-additional-results}), showing a similar pattern. Note that we also include results on the large-scale Manhattan2668 in Table~\ref{tbl:additional-results} (averaged over three different types of traffic flow), where gains of HALO are more significant (ATT 9.1\%, ADT 7.5\% better). Overall, these trends are visualized in Figure~\ref{fig:gain-vs-size}, where clearly HALO’s benefit grows with network size.

\noindent\textbf{Performance on Large-Scale Road Networks.}
We also consider a large-scale Manhattan road network (2,668 intersections) with three realistic traffic patterns: Peak Transition, Adverse Weather and Holiday Rush. As shown in Table~\ref{tbl:manhattan}, HALO achieves the lowest ATT/ADT in all cases: 690.88/549.02s (Peak), 913.84/649.71s (Weather), and 980.24/598.65s (Holiday). Comparing to the strongest baselines, this yields about 5.7\% (Peak) and 6.8\% (Holiday) ATT gains, with a smaller but consistent 1.8\% gain under Weather; ADT improves by 3.3\% (Peak), 2.6\% (Weather), and 5.0\% (Holiday). We attribute the larger deltas here to the non-stationary, non-local effects of real-world traffic in large road networks - formation/breakup of platoons, spillback propagation, corridor re-synchronization - where HALO’s outperforms the baselines due to (i) its hieratical design, where global guidance signal aligns local agents with corridor-level priorities (progression/load-balancing) - even more crucial in such a large road network; and (ii) the adversarial goal setting, which prevents trivial or overly conservative targets, discouraging locally stable but globally suboptimal equilibria.

\begin{table}[ht]
\begin{center}
\centering
\vspace{0pt}
\renewcommand{\arraystretch}{1.2}
\caption{Computational cost at inference time.}
\vspace{-1\baselineskip}
\label{flops}

\small
\begin{tabular}{>{\centering\arraybackslash}m{2cm}>{\centering\arraybackslash}m{2.2cm}>{\centering\arraybackslash}m{0.8cm}>{\centering\arraybackslash}m{0.8cm}>{\centering\arraybackslash}m{1cm}}

\noalign{\hrule height 1.5pt}

\cellcolor{gray!20}\textbf{Scenario} & 
\cellcolor{gray!20}\textbf{Metric ( /step )} & 
\cellcolor{gray!20}\textbf{HALO} & 
\cellcolor{gray!20}\textbf{GESA} & 
\cellcolor{gray!20}\textbf{X-Light} \\

\hline

 & FLOPs ($\times10^6$) & 25.2 & 0.96 & 2.53 \\
\multirow{-2}{1.8cm}{\centering Grid4×4} & Latency (ms) & 0.43 & 0.31 & 1.23 \\
\cellcolor{black!5} & \cellcolor{black!5}FLOPs ($\times10^6$) & \cellcolor{black!5}707 & \cellcolor{black!5}160 & \cellcolor{black!5}422 \\
\cellcolor{black!5}\multirow{-2}{1.8cm}{\centering Manhattan2668} & \cellcolor{black!5} Latency (ms) & \cellcolor{black!5}6.4 & \cellcolor{black!5}4.9 & \cellcolor{black!5}9.8 \\

\noalign{\hrule height 1.5pt}

\end{tabular}
\end{center}
\vspace{-1\baselineskip}
\end{table}


\noindent\textbf{Computational Efficiency.}
It is well known that stronger global coordination often comes at the cost of higher computation at inference time. We measure FLOPs and wall-clock time of the proposed HALO, and two strongest-performing baselines (GESA~\cite{jiang2024general} and X-Light~\cite{DBLP:conf/ijcai/Jiang00XRLM024}), for both small (Grid4$\times$4) and large (Manhattan2668) road networks (implementation details provided in Appendix~\ref{app:exp-efficiency} ). As shown in Table~\ref{flops}, per control step HALO runs in 0.43ms on Grid4$\times$4 (vs. 0.31ms for GESA and 1.23ms for X-Light) and 6.4ms on Manhattan2668 (vs. 4.9ms and 9.8ms) - well within a typical 10s ATSC control interval. Despite higher FLOPs, HALO’s run-time latency remains low because the global guidance policy executes once per 10 steps and thus can be amortized. Beyond latency, HALO’s coordination cost scales as $O(N)$ - a single broadcast guidance vector appended to each local observation - whereas dense neighbor messaging schemes can approach $O(N^2)$ in the worst case. Therefore the hierarchical design of HALO preserves global awareness of the city-wide traffic features without incurring quadratic cost, while allowing local policies to make informed decisions given the global context. Overall, HALO satisfies real-time constraints while delivering the strongest performance, making the modest compute overhead a worthwhile trade-off.

\begin{figure}[tbp]
    \begin{subcaptionbox}{\label{fig:ablations}}[0.49\linewidth]
        {\includegraphics[width=\linewidth]{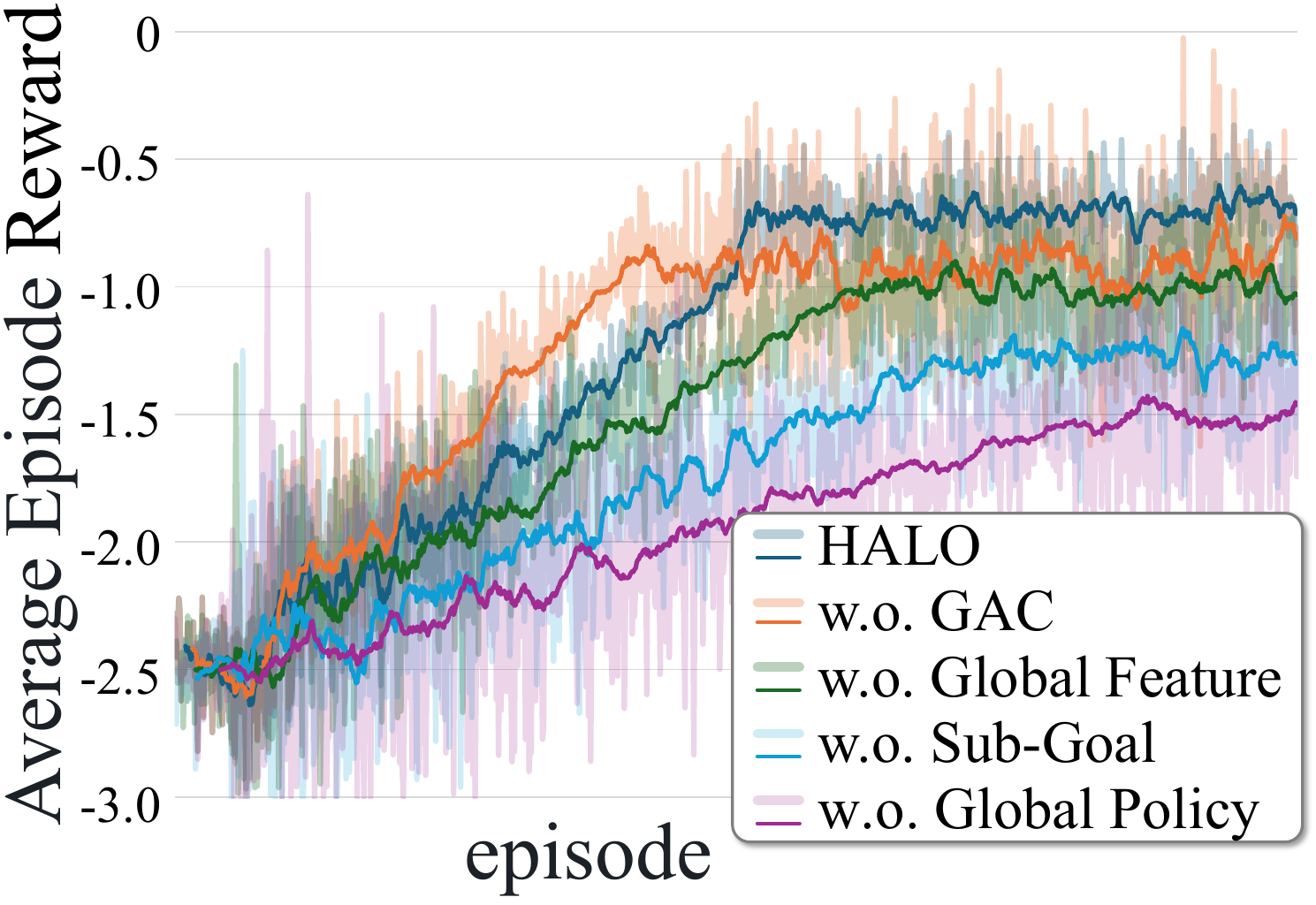}}
    \end{subcaptionbox}
    \hfill
    \begin{subcaptionbox}{\label{fig:pretraining}}[0.49\linewidth]
        {\includegraphics[width=\linewidth]{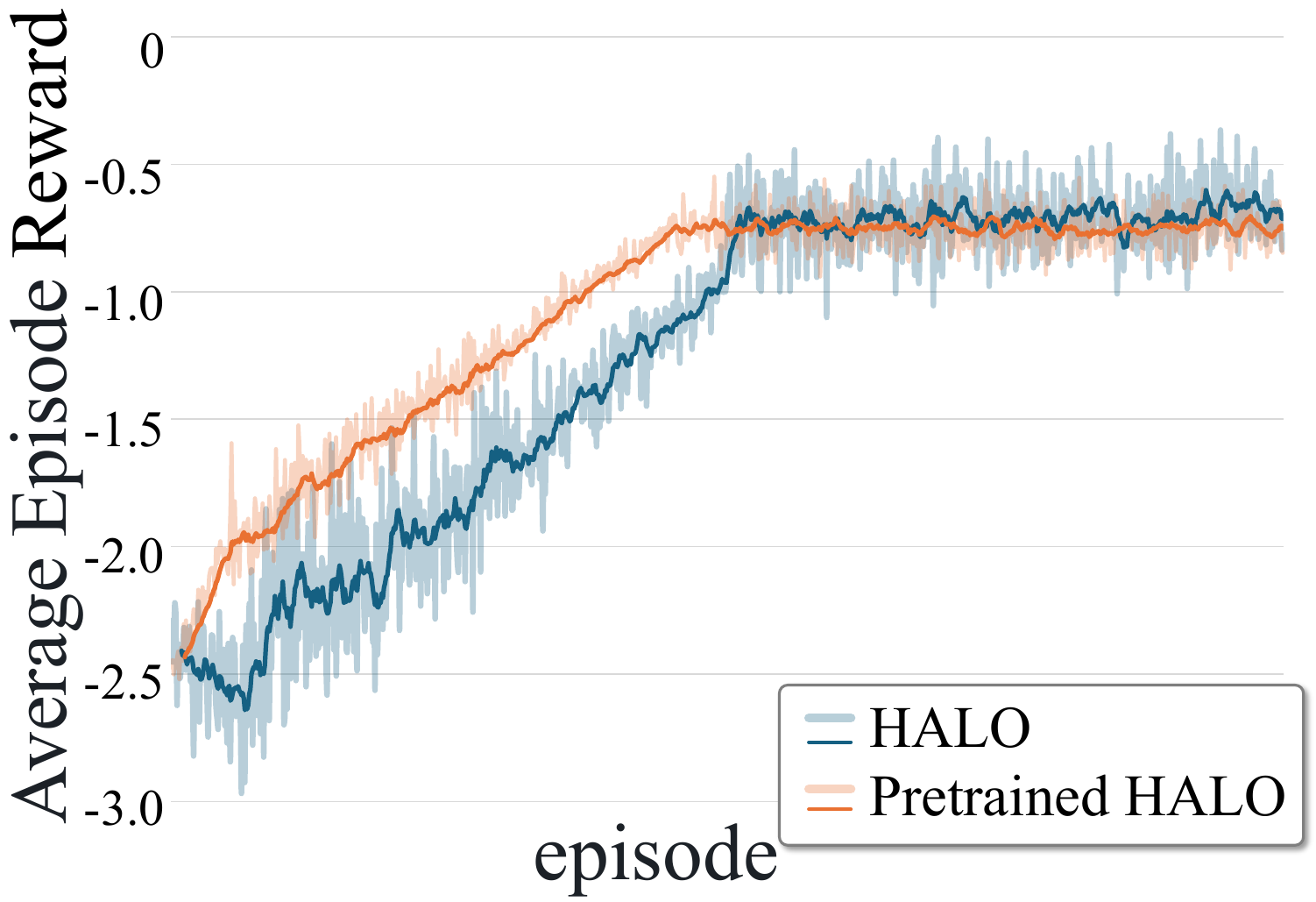}}
    \end{subcaptionbox}
    \vspace{-1\baselineskip}
    \caption{Episode–reward curves of HALO variants (a) with different components removed; and (b) with and without pretraining of the global guidance policy.}
    \Description{
    This figure presents two training analysis experiments using episode-reward curves.
    (a) shows an ablation study comparing the complete HALO model against four variants with progressively removed components: without Graph Attention-based Coordination (w.o. GAC), without Global Feature, without Sub-Goal, and without the entire global policy (w.o. global policy).
    The complete HALO model exhibits the fastest convergence and highest final reward, demonstrating stable performance throughout training.
    Removing GAC causes slight performance degradation with increased oscillations after convergence, indicating reduced local coordination stability.
    Excluding Global Feature or Sub-Goal leads to more significant performance drops, with slower convergence and lower final rewards, highlighting the importance of global context and hierarchical guidance.
    The most severe degradation occurs when removing the entire global policy, resulting in substantially lower rewards and slower learning.
    (b) illustrates the impact of pretraining the global guidance policy.
    The pretrained version shows a smoother learning curve without the initial reward drop observed in the baseline HALO, indicating that pretraining helps the global policy generate more informative sub-goals early in training.
    Although the learning slope becomes gentler in later stages, the pretrained model achieves slightly faster convergence while maintaining comparable final performance to the non-pretrained version.
    Together, these curves validate the necessity of each component in HALO's hierarchical architecture and demonstrate the benefit of global policy pretraining for training efficiency.
    }
\end{figure}

\noindent \textbf{Ablations.}
We further investigate the contributions of several core components of HALO by progressively removing them - resulting in different vanilla versions of the proposed approach - and show the Episode–Reward curves in Figure~\ref{fig:ablations}. We see that dropping GAC (neighbor Graph Attention) yields a small performance dip but noticeably more post-convergence oscillation, indicating reduced local-coordination stability. Removing the global observations (the guidance signal $F_g$) causes a larger decline and slower, less stable convergence, consistent with loss of long-range dependencies. Disabling adversarial goal setting but keeping the guidance signal $F_g$ also degrades performance: guidance becomes uninformative, learning slows, and final rewards fall substantially. The worst outcome arises when the entire global guidance mechanism is removed, eliminating both global modeling and goal generation. We also study the performance of HALO when pretraining of the global guidance policy is affordable. As shown in Figure~\ref{fig:pretraining}, pretraining the global guidance policy smooths the early learning phase and slightly accelerates convergence, though final performance is similar. In sum, these results confirm that global guidance and adversarial goal setting are the primary drivers of HALO’s gains, while GAC mainly stabilizes local interactions - this aligns with the larger improvements observed on the large-scale Manhattan2668 scenario with realistic traffic patterns, where non-stationary, non-local effects dominate.

\section{Conclusion}
\label{sec:conclusion}

In this paper, we introduce HALO, a novel hierarchical reinforcement learning framework designed to resolve the critical scalability-coordination trade-off in large-scale Adaptive Traffic Signal Control. By decoupling network-level planning from local execution, HALO effectively balances global strategic coordination with local responsiveness. In particular, the global guidance policy, powered by a Transformer-LSTM architecture, captures city-wide spatiotemporal dynamics to provide all intersections with a unified context, while the proposed adversarial goal-setting mechanism fosters robust alignment and improves learning efficiency. Extensive experiments demonstrate HALO's consistent outperformance of state-of-the-art methods, with its advantages becoming more pronounced as network scale and traffic complexity increase. On a large-scale Manhattan network with 2,668 intersections under realistic traffic conditions, HALO achieved significant improvements of up to 6.8\% in travel time and 5.0\% in delay. These results showcase that HALO's hierarchical design with explicit global guidance is a highly effective and computationally feasible solution for managing traffic in the complex, large-scale WoT environments of modern smart cities.

\begin{acks}
This work was supported by the Engineering and Physical Sciences Research Council (EPSRC) under Grant EP/Y016289/1.
\end{acks}

\bibliographystyle{ACM-Reference-Format}
\balance
\bibliography{reference}

\appendix

\section{Additional Details of HALO Design}
\label{app:additional-design-details}

In designing the observation space for HALO, we carefully select a set of observations that capture the real-time traffic state at each intersection. These observations reflect critical aspects of traffic flow, congestion, and signal efficiency, enabling the model to respond dynamically to varying traffic conditions. Table~\ref{tbl:observations} presents a detailed summary of each observation, including its semantic meaning and the method used for computation. In addition, the regional state includes three components: the "stop\_car\_num" and "waiting\_time" within the region, and the spatial coordinates of the subregion.


\begin{table*}[htbp]
\begin{center}
\centering
\vspace{0pt}
\renewcommand{\arraystretch}{1.2}
\vspace{-0.5\baselineskip}
\caption{Detailed descriptions and computation of observations}
\vspace{-1\baselineskip}
\label{tbl:observations}

\small
\resizebox{\textwidth}{!}{
\begin{tabular}{>{\centering\arraybackslash}m{2cm}>{\centering\arraybackslash}m{7.5cm}>{\centering\arraybackslash}m{7cm}}

\noalign{\hrule height 1.5pt}

\cellcolor{gray!20}\textbf{Observation} & 
\cellcolor{gray!20}\textbf{Meaning} & 
\cellcolor{gray!20}\textbf{Computation} \\

\hline

car\_num & Total number of vehicles on a lane, indicating the overall traffic density at the intersection. & The sum of all vehicles on each lane. \\
\cellcolor{black!5}queue\_length & \cellcolor{black!5}Length of the queue on incoming lanes, reflecting congestion or signal-induced queuing. & \cellcolor{black!5}Queue length in meters; represented as an N-dimensional vector, where N is the number of incoming lanes. \\
occupancy & Lane occupancy ratio, indicating the proportion of the lane occupied by vehicles. & The number of vehicles divided by lane capacity, per lane; N-dimensional. \\
\cellcolor{black!5}flow & \cellcolor{black!5}Number of vehicles passing through per unit time, measuring traffic throughput. & \cellcolor{black!5}The total number of vehicles passing through the intersection per unit time. \\
stop\_car\_num & Number of stationary vehicles on a lane (e.g., stopped due to red lights or congestion). & The number of vehicles with speed $< 0.1 m/s$, normalized by lane capacity; N-dimensional. \\
\cellcolor{black!5}waiting\_time & \cellcolor{black!5}Waiting time of the first vehicle on each incoming lane, reflecting delay at the intersection. & \cellcolor{black!5}Sum of waiting times for the first vehicle on each incoming lane; N-dimensional. \\
average\_speed & Average speed of vehicles passing through the intersection, reflecting potential delays due to signal changes. & The average speed of all vehicles passing through. \\
\cellcolor{black!5}pressure & \cellcolor{black!5}Traffic pressure in a specific direction, reflecting imbalance between demand and discharge capacity. & \cellcolor{black!5}The difference in vehicle count between incoming and corresponding outgoing lanes. \\
delay\_time & The time loss due to the vehicle not reaching its expected speed. & The difference between the ideal travel time and the actual travel time. \\

\noalign{\hrule height 1.5pt}

\end{tabular}}
\end{center}
\vspace{-1\baselineskip}
\end{table*}


\section{Implementation Details}
\label{app:impl-details}

\subsection{Hyper-parameter Settings}
\label{app:impl-hyperparams}

Table~\ref{tbl:params} summarizes the key hyperparameters used in our experiments. When dividing the network into subregions using a grid-based approach, the number of intersections per region is generally kept within the order of $10^1$ to ensure training efficiency. For example, in the Manhattan2668 scenario, which contains a total of 2668 intersections and features a long and narrow layout, we partition the network into a 4×15 grid, resulting in 60 regions with an average of approximately 45 intersections per region.


\begin{table}[h]
\begin{center}
\centering
\vspace{0pt}
\renewcommand{\arraystretch}{1.2}
\caption{Key hyperparameters used in experiments}
\vspace{-1\baselineskip}
\label{tbl:params}

\small
\begin{tabular}{>{\centering\arraybackslash}m{2.5cm}>{\centering\arraybackslash}m{0.8cm}|>{\centering\arraybackslash}m{2.5cm}>{\centering\arraybackslash}m{0.8cm}}

\noalign{\hrule height 1.5pt}

\cellcolor{gray!20}\textbf{Description} & 
\cellcolor{gray!20}\textbf{Value} & 
\cellcolor{gray!20}\textbf{Description} & 
\cellcolor{gray!20}\textbf{Value} \\

\hline

Total Steps & $380000$ & Hidden Layer Size & $64$ \\
\cellcolor{black!5}Clipping Parameter & \cellcolor{black!5}$0.2$ & \cellcolor{black!5}Output Layer Gain & \cellcolor{black!5}$0.01$ \\
Entropy Coefficient & $0.01$ & Number of Actions & $8$ \\
\cellcolor{black!5}Value Loss Coefficient & \cellcolor{black!5}$1.0$ & \cellcolor{black!5}SUMO Start Port & \cellcolor{black!5}$15000$ \\
Learning Rate & $3e-4$ & Episode Duration & $3600$s \\
\cellcolor{black!5}Discount Factor & \cellcolor{black!5}$0.99$ & \cellcolor{black!5}Yellow Light Duration  & \cellcolor{black!5}$5$s \\
GAE Lambda & $0.95$ & Control Interval  & $10$s \\
\cellcolor{black!5}Maximum Gradient Norm & \cellcolor{black!5}$10.0$ & \cellcolor{black!5}Observation Dimension & \cellcolor{black!5}$66$ \\

\noalign{\hrule height 1.5pt}

\end{tabular}
\end{center}
\vspace{-1\baselineskip}
\end{table}


To analyze the sensitivity of the model’s performance to the network partitioning strategy, we conducted a grid-size hyperparameter analysis, and the results are presented in Table~\ref{tbl:GridSize}. The table reports the number of grid cells, the average number of intersections per cell, and the relative performance in ATT and ADT compared with the $4\times 15$ grid configuration used in our main experiments.


\begin{table}[ht]
\begin{center}
\centering
\vspace{0pt}
\renewcommand{\arraystretch}{1.2}
\caption{Grid size hyperparameter analysis}
\vspace{-1\baselineskip}
\label{tbl:GridSize}

\small
\begin{tabular}{>{\centering\arraybackslash}m{0.6cm}>{\centering\arraybackslash}m{0.85cm}>{\centering\arraybackslash}m{1.8cm}>{\centering\arraybackslash}m{1.7cm}>{\centering\arraybackslash}m{1.7cm}}

\noalign{\hrule height 1.5pt}

\cellcolor{gray!20}\textbf{Grid} & 
\cellcolor{gray!20}\textbf{\# Cells} & 
\cellcolor{gray!20}\textbf{Intersections /Cell} & 
\cellcolor{gray!20}\textbf{$ \Delta $ATT vs. 4×15} & 
\cellcolor{gray!20}\textbf{$ \Delta $ADT vs. 4×15} \\

\hline

2×2 & 4 & 667 & +17\% & +16\% \\
\cellcolor{black!5}3×3 & \cellcolor{black!5}9 & \cellcolor{black!5}296 & \cellcolor{black!5}+8\% & \cellcolor{black!5}+7\% \\
2×8 & 16 & 167 & +0.9\% & +0.7\% \\
\cellcolor{black!5}3×12 & \cellcolor{black!5}36 & \cellcolor{black!5}74 & \cellcolor{black!5}<0.1\% & \cellcolor{black!5}<0.1\% \\
\rowcolor{green!10}4×15 & 60 & 45 & 0 & 0 \\
\cellcolor{black!5}6×25 & \cellcolor{black!5}150 & \cellcolor{black!5}18 & \cellcolor{black!5}+0.2\% & \cellcolor{black!5}-0.1\% \\

\noalign{\hrule height 1.5pt}

\end{tabular}
\end{center}
\vspace{-1\baselineskip}
\end{table}


As illustrated in Table~\ref{tbl:GridSize}, the model's performance remains stable across most partitioning strategies, with ATT and ADT nearly unchanged from 16 to 150 intersections per subregion. This robustness is due to the hierarchical design: the global guidance policy operates on compact, aggregated features that preserve key traffic patterns as long as the grid is not too coarse or too fine. The final 4×15 grid design for Manhattan was chosen based on the city’s spatial proportions, as it provides balanced spatial resolution along both vertical and horizontal directions.

\subsection{Training Details}
\label{app:impl-training-details}
We train our model using Adam optimizer on NVIDIA GeForce RTX 4090 for around 19 hours and test it on the same GPU.
Given the inherent challenges of adversarial training and multi-agent RL, maintaining training stability is crucial for the success of HALO. Several key training techniques are implemented to ensure robust convergence and prevent common training failures:

\begin{itemize}

\item \textbf{Two-time-scale updates}: The global guidance policy is updated once every ten local intersection policy steps. This approach decouples the adversarial feedback loop, effectively reducing oscillations and improving training stability.

\item \textbf{Stop-gradient on global targets}: Stop-gradient operations are applied to global targets ($G^t$) when updating the local intersection policy. This technique breaks the backward gradient flow from local intersection policy to global guidance policy, alleviating non-stationarity while preserving the guiding signal.

\item \textbf{Limited neighborhood in Graph Attention}: In the Graph Attention module, only the four nearest neighbors are retained in the adjacency matrix. This design choice reduces long-range message noise, lowers variance, and eliminates divergence issues observed with full-scale graphs.

\item \textbf{Gradient clipping and reward normalization}: Standard techniques such as gradient clipping and reward normalization are adopted, as omitting them led to occasional NaNs during early experiments.

\item \textbf{Adversarial mechanism for optimization guidance}: The adversarial training mechanism itself contributes to stability by providing carefully selected sub-goals that offer clearer optimization directions for multi-agent models, helping address optimization instability caused by sparse rewards.
\end{itemize}

\section{Experimental Details}
\label{app:exp-details}
\subsection{Visualization of Experimental Scenarios}
\label{app:exp-vis}
\begin{figure}[htbp]
\centerline{\includegraphics[scale=0.11]{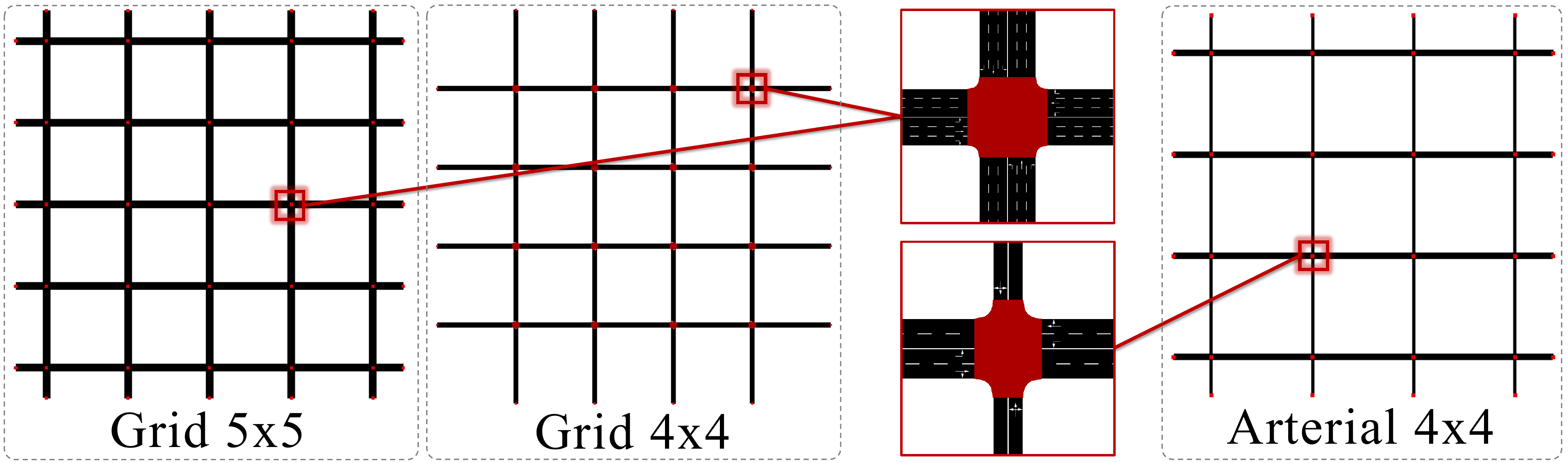}}
\vspace{-1\baselineskip}
\caption{Synthetic scenarios with sample intersections.}
\vspace{-1\baselineskip}
\label{map1}
\Description{This figure displays three synthetic traffic network scenarios used for controlled experimentation in traffic signal control research.
The left panel shows the Grid 4 by 4 scenario, consisting of 16 intersections arranged in a regular grid pattern with uniform spacing, representing a typical urban block structure.
The middle panel presents the Arterial 4 by 4 scenario, also containing 16 intersections but arranged along a main arterial road with connecting side streets, simulating a common urban corridor configuration.
The right panel illustrates the Grid 5 by 5 scenario, a larger grid network with 25 intersections providing increased complexity for scalability testing.
Each scenario includes a zoomed-in view of sample intersections showing the detailed lane configurations, signal phases, and typical four-leg intersection geometry with incoming and outgoing lanes for different movement directions including left turns, through movements, and right turns.}
\end{figure}

\begin{figure}[htbp]
\centerline{\includegraphics[scale=0.12]{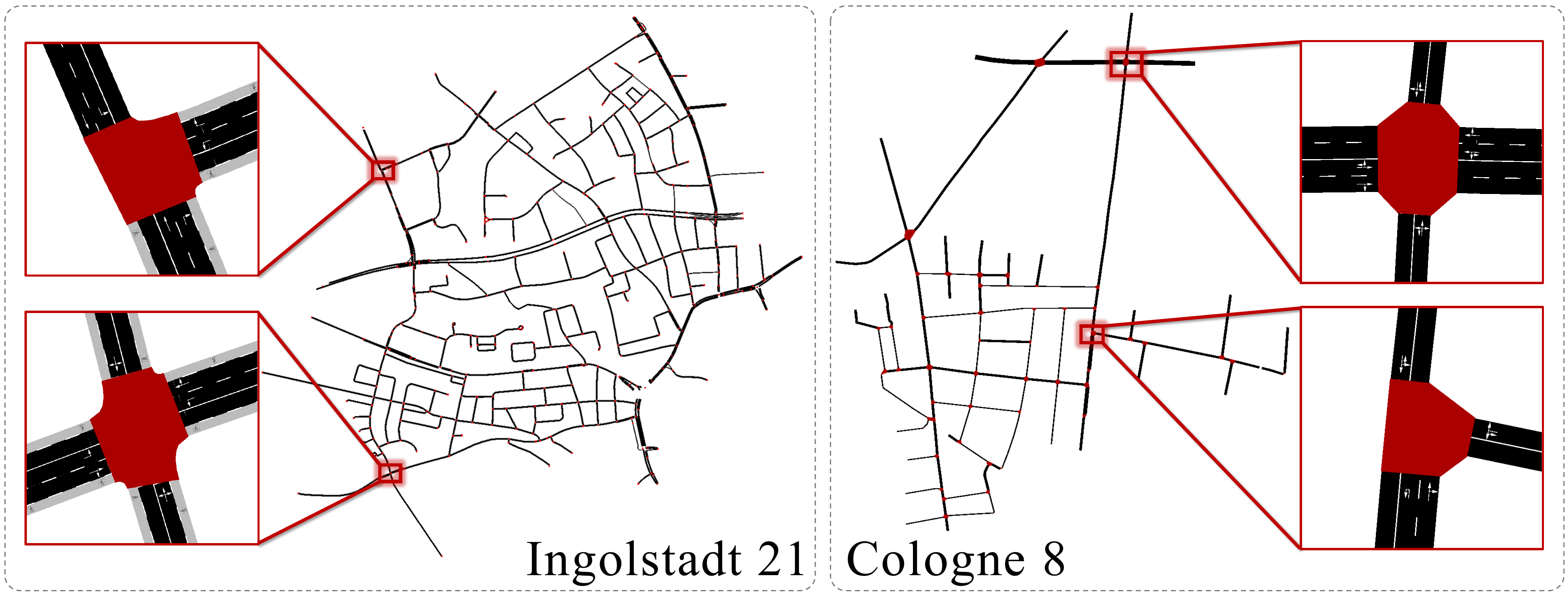}}
\vspace{-1\baselineskip}
\caption{German real-world scenarios with sample intersections}
\vspace{-1\baselineskip}
\label{map2}
\Description{This figure presents two real-world traffic network scenarios based on German cities, providing realistic test environments for adaptive traffic signal control algorithms.
The left panel shows the Cologne 8 scenario, extracted from the road network of Cologne, Germany, containing 8 signalized intersections with irregular spacing and heterogeneous intersection geometries that reflect actual urban infrastructure.
This scenario captures morning peak traffic flow patterns recorded from real-world data, including variations in lane configurations and non-uniform traffic demands across different approaches.
The right panel displays the Ingolstadt 21 scenario, derived from the city of Ingolstadt, Germany, featuring 21 intersections with more complex network topology including varied intersection types, multiple merging points, and diverse road hierarchies.
Like Cologne 8, this scenario uses real traffic flow data collected during morning peak hours, representing authentic congestion patterns and driver behaviors.
Both scenarios include zoomed-in views of representative intersections illustrating the actual lane layouts, signal configurations, and geometric characteristics found in European urban environments.}
\end{figure}

\begin{figure*}[htbp]
\centerline{\includegraphics[scale=0.356]{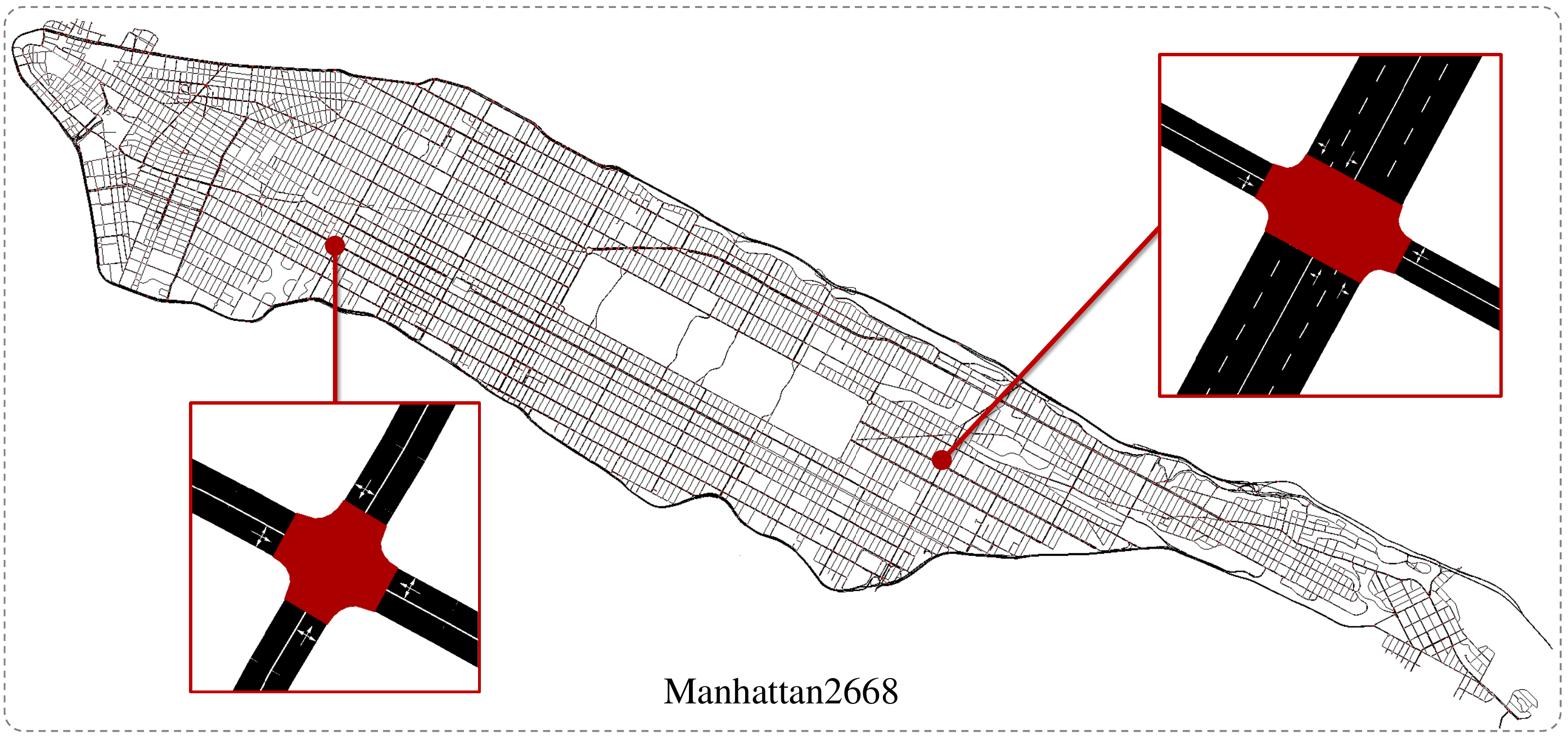}}
\caption{Manhattan real-world scenario with sample intersections}
\vspace{-1\baselineskip}
\label{map3}
\Description{This figure illustrates the large-scale Manhattan 2668 real-world traffic network scenario, representing the most challenging and realistic test environment used in this study.
The main panel shows the complete road network of Manhattan, New York City, containing 2668 signalized intersections extracted from OpenStreetMap data and refined with actual intersection details.
The network exhibits the characteristic elongated grid structure of Manhattan Island, spanning from lower Manhattan to the northern tip, with the dense urban grid punctuated by major arterials and irregular blocks near parks and waterfront areas.
This scenario represents a true Web-of-Things scale urban traffic system with thousands of interconnected sensing and control nodes.
Three zoomed-in inset panels display sample intersections from different regions of the network, showcasing the diversity of intersection geometries, lane configurations, and signal phase arrangements found across Manhattan.
These samples include standard grid intersections with regular four-leg geometry, complex multi-way intersections near major avenues, and irregular junctions influenced by diagonal streets and urban planning constraints.}
\end{figure*}

Figure~\ref{map1}, Figure~\ref{map2}, and Figure~\ref{map3} illustrate representative scenarios used in our experiments.

\subsection{Details on Computational Efficiency}
\label{app:exp-efficiency}
We use the official code repositories provided by GESA~\cite{jiang2024general} and X-Light~\cite{DBLP:conf/ijcai/Jiang00XRLM024}. 
All experiments are conducted on an NVIDIA GeForce RTX 4090 GPU using PyTorch with CUDA acceleration. For each method and scenario, we measure wall-clock inference time using PyTorch's \texttt{torch.cuda.Event}, which provides microsecond-precision timing on GPU. The measurement protocol consists of the following: (i) a warm-up phase of 100 control steps to eliminate GPU kernel compilation overhead and ensure stable GPU clock frequencies; (ii) a measurement phase of 1000 forward passes, recording the time required to compute actions for all intersections in each control step; and (iii) aggregation, where we compute both the mean latency and 95th percentile latency to account for potential outliers. The reported values in Table~\ref{flops} represent the mean latency per control step. FLOPs are calculated using the standard profiling tool \texttt{fvcore.nn.FlopCountAnalysis}, which counts multiply-add operations in the forward pass, with input dimensions matching the actual observation sizes used in each scenario.

\subsection{Additional Results}
\label{app:exp-additional-results}
This section presents additional performance comparisons on Colo-gne8, Grid4×4, and Manhattan2668 scenarios to complement the results shown in Section~\ref{exp:results} of the main text. Note that the results of Manhattan2668 in Table~\ref{tbl:additional-results} are calculated as the average over the three traffic flow settings (Peak Transition, Adverse Weather, and Holiday Rush) as reported in Table~\ref{tbl:manhattan}.


\begin{table}[htbp]
\centering
\renewcommand{\arraystretch}{1.2}

\caption{Performance of competing approaches on both standard scenarios (Cologne8, Grid4×4) and the large-scale Manhattan2668. Best in bold red, second best in blue, and improvements shown by \better{\%}.}
\vspace{-1\baselineskip}
\label{tbl:additional-results}

\small
\begin{tabular}{>{\centering\arraybackslash}m{1.4cm}>{\centering\arraybackslash}m{1.3cm}>{\centering\arraybackslash}m{1.7cm}>{\centering\arraybackslash}m{2cm}}

\noalign{\hrule height 1.5pt}

\cellcolor{gray!20}\textbf{Methods} & 
\cellcolor{gray!20}\textbf{Cologne8} & 
\cellcolor{gray!20}\textbf{Grid4×4} & 
\cellcolor{gray!20}\textbf{Manhattan2668} \\

\hline
\multicolumn{4}{l}{\cellcolor{gray!10}\textbf{Avg. Trip Time (seconds) $\downarrow$}} \\
\hline

FTC & 124.40\tiny{±1.99} & 206.68\tiny{±0.54} & 1844.56\tiny{±9.13} \\
\cellcolor{black!5} MaxPressure & \cellcolor{black!5} 95.96\tiny{±1.11} & \cellcolor{black!5} 175.97\tiny{±0.70} & \cellcolor{black!5} 1441.23\tiny{±8.26} \\
CoLight & \second{ 89.72\tiny{±0.00}} & 163.52\tiny{±0.00} & 1138.66\tiny{±0.00} \\
\cellcolor{black!5} MPLight & \cellcolor{black!5} 98.44\tiny{±0.62} & \cellcolor{black!5} 179.51\tiny{±0.95} & \cellcolor{black!5} 1374.73\tiny{±18.46} \\
MetaLight & 91.57\tiny{±0.75} & 169.21\tiny{±1.26} & 1029.04\tiny{±66.54} \\
\cellcolor{black!5} IPPO & \cellcolor{black!5} 90.87\tiny{±0.40} & \cellcolor{black!5} 167.62\tiny{±2.42} & \cellcolor{black!5} 1293.32\tiny{±71.79} \\
rMAPPO & 97.68\tiny{±2.03} & 164.96\tiny{±1.87} & 1348.98\tiny{±24.61} \\
\cellcolor{black!5} MetaGAT & \cellcolor{black!5} 96.74\tiny{±0.00} & \cellcolor{black!5} 165.23\tiny{±0.00} & \cellcolor{black!5} 974.93\tiny{±0.00} \\
GESA & 90.59\tiny{±0.74} & 161.33\tiny{±1.34} & \second{ 948.09\tiny{±14.63}} \\
\cellcolor{black!5} CoSLight & \cellcolor{black!5} 90.46\tiny{±0.55} & \cellcolor{black!5} \second{ 159.11\tiny{±3.12}} & \cellcolor{black!5} 1053.90\tiny{±26.04} \\
X-Light & \best{88.55\tiny{±0.00}} & 162.47\tiny{±0.00} & 999.61\tiny{±0.00} \\
\rowcolor{green!10} \textbf{HALO} & 90.83\tiny{±0.34} & \best{159.07\tiny{±0.97}}\better{0.0\%} & \best{861.65\tiny{±14.78}}\better{9.1\%} \\

\hline
\multicolumn{4}{l}{\cellcolor{gray!10}\textbf{Avg. Delay Time (seconds) $\downarrow$}} \\
\hline

FTC & 62.38\tiny{±2.95} & 94.64\tiny{±0.43} & 1477.97\tiny{±24.79} \\
\cellcolor{black!5} MaxPressure & \cellcolor{black!5} 31.93\tiny{±1.07} & \cellcolor{black!5} 64.01\tiny{±0.71} & \cellcolor{black!5} 1067.19\tiny{±14.47} \\
CoLight & 25.56\tiny{±0.00} & 51.58\tiny{±0.00} & 763.42\tiny{±0.00} \\
\cellcolor{black!5} MPLight & \cellcolor{black!5} 34.38\tiny{±0.63} & \cellcolor{black!5} 67.52\tiny{±0.97} & \cellcolor{black!5} 864.85\tiny{±8.26} \\
MetaLight & 27.61\tiny{±0.78} & 57.56\tiny{±0.76} & 688.79\tiny{±36.1} \\
\cellcolor{black!5} IPPO & \cellcolor{black!5} 26.82\tiny{±0.43} & \cellcolor{black!5} 56.38\tiny{±1.46} & \cellcolor{black!5} 874.35\tiny{±13.81} \\
rMAPPO & 33.37\tiny{±1.97} & 53.65\tiny{±1.00} & 889.84\tiny{±23.69} \\
\cellcolor{black!5} MetaGAT & \cellcolor{black!5} 26.85\tiny{±0.00} & \cellcolor{black!5} 50.32\tiny{±0.00} & \cellcolor{black!5} 733.22\tiny{±0.00} \\
GESA & 26.50\tiny{±0.87} & 49.60\tiny{±0.71} & 675.64\tiny{±7.17} \\
\cellcolor{black!5} CoSLight & \cellcolor{black!5} \second{ 24.79\tiny{±1.23}} & \cellcolor{black!5} \best{46.11\tiny{±1.21}} & \cellcolor{black!5} 659.04\tiny{±10.78} \\
X-Light & \best{24.31\tiny{±0.00}} & 50.27\tiny{±0.00} & \second{ 647.73\tiny{±0.00}} \\
\rowcolor{green!10} \textbf{HALO} & { 24.81\tiny{±0.93}} & \second{ 48.25\tiny{±0.53}} & \best{599.13\tiny{±5.60}}\better{7.5\%} \\

\noalign{\hrule height 1.5pt}

\end{tabular}
\vspace{-1\baselineskip}
\end{table}


\section{Limitations and Future Work}

While HALO demonstrates strong scalability and performance in large-scale adaptive traffic signal control, several limitations remain that suggest meaningful directions for future work.

\paragraph{Global observability and communication assumptions.} HALO assumes that network-level information can be reliably aggregated and broadcast by the global guidance policy. In real-world city-scale systems, communication delays, packet loss, or sensor failures may lead to partial or stale global observations. While local intersection policies can still operate based on local states, degraded global guidance may reduce coordination effectiveness. Future work could explicitly model partial observability or uncertainty in the global policy.

\paragraph{Robustness to delayed or imperfect global guidance.}
The current framework assumes synchronized delivery of global guidance to all intersections. In practice, guidance may arrive with heterogeneous delays or be temporarily unavailable. Incorporating delay-aware conditioning or adaptive reliance on global signals could improve robustness. Evaluating HALO under controlled delay and packet-loss settings is a natural next step.

\paragraph{Sim-to-real and deployment challenges.}
HALO is evaluated in simulation and does not fully capture real-world issues such as noisy sensors, detector failures, or unexpected incidents. Although the hierarchical design and adversarial goal setting improve robustness under non-stationary traffic patterns, additional techniques such as domain randomization or online adaptation would be required for deployment. Testing HALO in real or semi-real traffic systems remains an important direction for future work.



\end{document}